\documentclass[10pt,twocolumn,letterpaper]{article}

\usepackage{cvpr}              % To produce the CAMERA-READY version
\usepackage[table]{xcolor}
\usepackage[normalem]{ulem}

\usepackage{todonotes}

\definecolor{sorcenrow}{rgb}{0.525, 0.796, 0.863}
\usepackage{multirow}
\usepackage{subcaption}
\usepackage{comment}

\usepackage{csquotes}
\usepackage{soul}

\usepackage{placeins}

\definecolor{cvprblue}{rgb}{0.21,0.49,0.74}
\usepackage[pagebackref,breaklinks,colorlinks,allcolors=cvprblue]{hyperref}

\def\paperID{*****} % *** Enter the Paper ID here
\def\confName{CVPR}
\def\confYear{2026}

\title{Learning from Semantic Dictionaries: Discriminative Codebook Contrastive Learning for Unified Visual Representation and Generation}

\author{
Imanol G. Estepa$^{1}$ \quad
Jesús M. Rodríguez-de-Vera$^{1}$ \quad
Bhalaji Nagarajan$^{2}$ \quad
Petia Radeva$^{1}$\\[0.5em]
$^{1}$Universitat de Barcelona, Barcelona, Spain\\
$^{2}$Barcelona Supercomputing Center (BSC), Barcelona, Spain\\ [0.5em]
{\tt\small estepa.gonzalez@ub.edu, j.molina.rdv@ub.edu, bhalaji.nagarajan@bsc.es, petia.radeva@ub.edu} \\
}

\begin{document}
\maketitle

% \todo[inline]{Experiments}
% \todo[i]{}

% \begin{abstract}
% The ABSTRACT is to be in fully justified italicized text, at the top of the left-hand column, below the author and affiliation information.
% Use the word ``Abstract'' as the title, in 12-point Times, boldface type, centered relative to the column, initially capitalized.
% The abstract is to be in 10-point, single-spaced type.
% Leave two blank lines after the Abstract, then begin the main text.
% Look at previous \confName abstracts to get a feel for style and length.
% \end{abstract}

\begin{abstract}

Discriminative and generative vision models excel in their respective domains but remain semantically misaligned, hindering progress toward unified visual learning. We introduce \textbf{LEASE} (LEArning from SEmantic Dictionaries), a self-supervised framework that bridges this gap using a paired generative–discriminative codebook design. LEASE operates entirely in a discrete token space produced through a one-time precomputation step, enabling efficient training without data augmentations, teacher models, or online tokenizers.
LEASE integrates two complementary objectives: a \textbf{masked token reconstruction loss} that captures fine-grained generative detail, and a \textbf{codebook contrast loss} that aligns encoder features with discriminative semantics via adaptive centroid weighting. This dual supervision yields a unified latent space that supports both high-quality generation and strong representation learning.
On ImageNet-1K, LEASE achieves state-of-the-art unified performance, outperforming prior VQGAN-based methods such as MAGE and Sorcen across linear probing, unconditional generation, few-shot learning, transfer, and robustness benchmarks. It also competes favorably with domain-specialized contrastive and generative models while surpassing previous MIM methods. The unsupervised LEASE model can also be extended to conditional generation by building upon its learned representations, proving competitive with specialized baselines.
Overall, LEASE provides an efficient and effective step toward general-purpose vision models that jointly understand and generate visual content. 
Code available at \url{https://github.com/ImaGonEs/LEASE}.

\end{abstract}    

\section{Introduction}
\label{sec:introduction}

% STORYLINE
% Vision Foundation models seem to rule; end with discriminative guidance for 
% Start with guidance, smoothly move to generation and mention tokenizers
% papers that initialize from VFMs -> train using VFMs as codebooks or train using Generative Codebooks -> Interesting parallellism again 
% while no apparent relation, we prove both can be useful

\begin{figure}[h]
    \centering
    %\begin{minipage}[c][120pt][c]{0.9\linewidth}
    
    \includegraphics[width=0.85\linewidth]{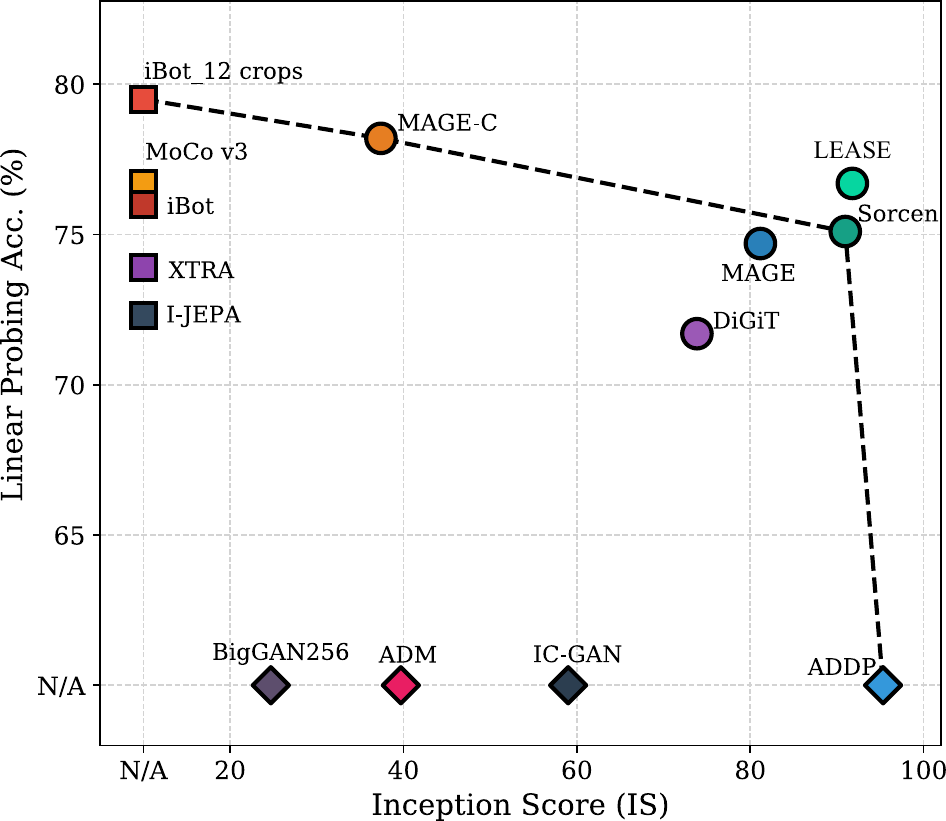}
    %\end{minipage}
\caption{
\textbf{Linear probing and class-unconditional generation performance.} 
% Each point represents a model. 
Our approach, \textit{LEASE}, advances the Unified SSL 
% state-of-the-art (SoTA) 
SoTA and defines a new Pareto-front for unified % hybrid 
models.
}
    \label{fig:pareto}
\end{figure}

Vision Foundation Models (VFM) have emerged as % are amongst the 
one of the most successful computer vision architectures \cite{zhou_ibot_2022, oquab_dinov2_2024, najdenkoska_tulip_2025, simeoni_dinov3_2025}. 
Trained largely on self-supervised or weakly supervised objectives \cite{caron_emerging_2021, radford_learning_2021}, they learn rich latent representations that support a wide range of discriminative tasks. However, these representations cannot be directly translated into high-fidelity visual generation \cite{li_mage_2023, li_mergevq_2025, tian_addp_2024}. 
Even Masked Image Modeling (MIM) methods such as SimMIM \cite{xie_simmim_2022} and MAE \cite{he_masked_2022}, that are explicitly optimized for reconstruction, often fail to produce detailed, realistic images when decoded from their learned latent representations. 

In contrast, % to VFMs, 
generative models including diffusion \cite{rombach_high-resolution_2022}, GAN-based architectures \cite{brock_large_2018} and Vector Quantization tokenizers \cite{esser_taming_2021, yu_vector-quantized_2021} capture rich low-level and mid-level visual details of training images to synthesize high-fidelity images, often guided by text or multimodal conditioning. 
%Vector quantization methods, such as VQGAN \cite{esser_taming_2021} and ViT-VQGAN \cite{yu_vector-quantized_2021}, explore the idea of encoding semantics relevant for generation into a discrete space via Codebooks. On inference, each image is mapped to a sequence of quantized tokens that summarize the most relevant semantics and, inversely, a model can be trained to generate these semantics and decode them back into an image.
%Many generative models even represent images through discrete codebooks, where each image is mapped to a sequence of quantized tokens that summarize the most relevant semantics. 
However, when their generative representations are applied for discriminative tasks, their performance drops significantly, mirroring the limitations observed in VFMs \cite{mukhopadhyay_text-free_2023,xiang_denoising_2023,chen_deconstructing_2024}. 
These features, while useful for image generation, lack global discriminative semantics common in VFMs. 
Although both paradigms aim at strong feature learning capabilities, they appear to \enquote{speak} fundamentally different semantic languages. 

%Although most VFMs are designed for recognition benchmarks \cite{simeoni_dinov3_2025, oquab_dinov2_2024}, these complex models could also contain sufficient information for image synthesis tasks in their complex latent spaces.
%In fact, recent \bhalaji{what?} works \cite{li_mergevq_2025, zhu_stabilize_2024, yu_representation_2024} leverage VFM features as guidance or conditioning for generative models, showing promising results.

%While different in nature, both semantic “languages” encode rich and complementary information that, if used jointly, can surpass the capabilities of models trained under a single paradigm. 
% While derived from 
Although trained with different objectives, semantic spaces of both models contain complementary information. 
Integrating them within a \emph{single model} yields stronger performance than relying on either paradigm independently.
% In fact, 
Recent unified and hybrid approaches leverage VFM features to guide % as guidance for 
generative models, improving generation while revealing emergent discriminative abilities \cite{li_mergevq_2025, zhu_stabilize_2024, yu_representation_2024}. 
Similarly, vector-quantization tokenizers like VQGAN \cite{esser_taming_2021} have enabled masked-token reconstruction frameworks that pretrain SSL models directly in token space, yielding strong performance in both discrimination and generation \cite{li_mage_2023, estepa_conjuring_2025}.
Other strategies including initializing generative models from pretrained VFM weights or distilling features from discriminative models, have % which also 
shown promising unified behavior \cite{li_mergevq_2025,  yu_representation_2024, zheng_vision_2025}. 
% However, these methods leverage distillation strategies as auxiliary model enhancements 
However, these methods treat unified performance as a byproduct of distillation or initialization
and do not address the underlying issue: % cause of the division:
the \textbf{semantic misalignment} between generative and discriminative representations. 
In this work, we directly target this misalignment and show that explicitly learning % training 
over % these two 
both semantic spaces % paradigms 
allows a single model to excel in both visual generation and discriminative representation learning.
%While different, both semantic languages encode rich information that, if managed symbiotically by a model, can outperform the single paradigm counterparts. Recent works, presented as unified or hybrid models, leverage VFM features as guidance or conditioning for generative models, showing improved generative results and an emergent discriminative capacity \cite{li_mergevq_2025, zhu_stabilize_2024, yu_representation_2024}. Similarly, vector quantization tokenizers such as VQGAN \cite{esser_taming_2021} enabled Masked Token Modelling frameworks, which leverage these generative tokenizers to pretrain self-supervised learning models to provide strong discriminative and generative performance \cite{li_mage_2023, estepa_conjuring_2025}. 
%Given these distinct “languages,” recent works have explored the idea of unifying discriminative and generative modeling, aiming to narrow the gap between the two paradigms. \bhalaji{We need to introduce the hybrid models before getting to initializing / distilling.}
%Approaches such as initializing generative models with pretrained VFM weights or distilling features from discriminative backbones have shown promising performance. Yet, they do not explore the root cause of the gap: the semantic misalignment between discriminative and generative representations. 
%In this work, we explicitly target this semantic gap and demonstrate that establishing a direct link between the two paradigms enables a model to excel in both generation and discrimination.

% \emph{LEA}rning from \emph{SE}mantic Dictionaries (LEASE) 
\textbf{\ul{LEA}}rning from \textbf{\ul{SE}}mantic Dictionaries (LEASE) introduces a unified framework that bridges generative and discriminative semantics through paired codebooks (referred as \enquote{dictionaries}). 
Each dictionary discretely encodes latent representations under a specific paradigm, either generation or discrimination, while providing % which provides 
complementary views of visual semantics. 
Instead of following VFM distillation strategies, LEASE jointly learns generative and discriminative perspectives by combining two objectives: a \textbf{masked token reconstruction loss} based on a generative dictionary and a \textbf{codebook contrastive loss} derived from a discriminative dictionary. 
The generative dictionary transforms images into sequences of discrete tokens, allowing LEASE to learn detailed visual information relevant for generative tasks just by reconstructing the masked input tokens. 
In parallel, the contrastive objective regularizes the latent space % these representations 
by aligning the features with the semantics encoded in the discriminative dictionary. 
Through this dual learning process, LEASE natively unites both \enquote{languages} within a shared latent space, yielding % enabling 
strong performance in both paradigms (\Cref{fig:pareto}). %, as shown in Figure \ref{fig:pareto}. 
Unlike previous unified % unification 
approaches, LEASE requires neither image augmentations nor additional frozen models, as it learns directly from the semantic differences encoded in its dictionaries. 
% The input tokens are extracted once before training, enabling an efficient and fully self-contained training.
All input tokens are precomputed once before training, resulting in an efficient, fully self-contained % , and semantically grounded unified 
learning framework.

\begin{figure}[t]
    \centering
    \includegraphics[width=\linewidth]{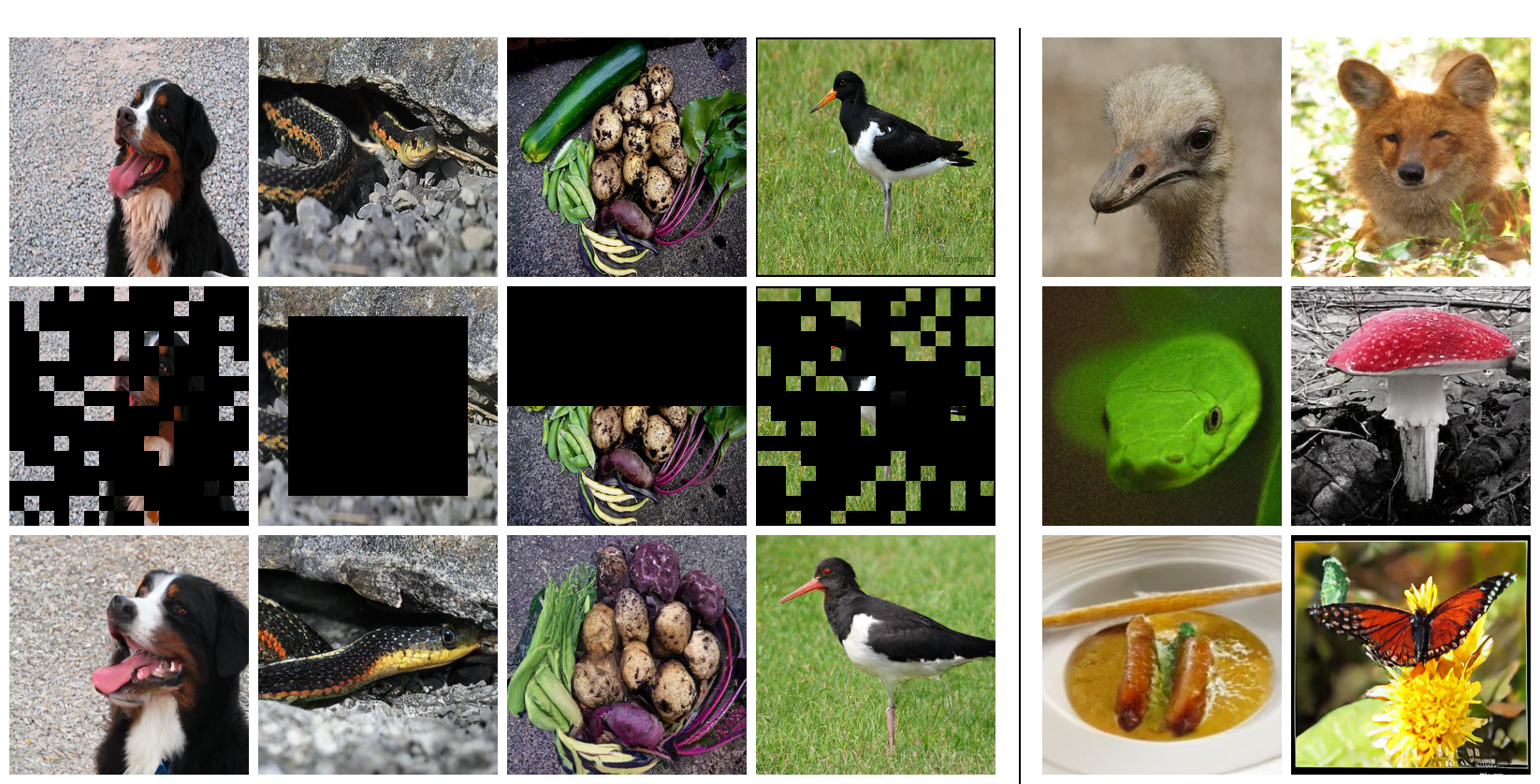}
    \caption{
    (Left) \textbf{Reconstruction} Samples with
    % using different masks.
    original images (Left-Top Row) and Masks (Left-Middle Row).
    (Right) \textbf{Conditional} (Col\# 1) and \textbf{Unconditional} (Col\# 2) generated samples.
    % (Left) \textbf{Reconstruction} using different masks. 
    % (Right) \textbf{Unconditionally and conditionally generated samples}, in order.
    }   
    \label{fig:placeholder}
\end{figure}

We evaluate LEASE across a diverse set of discriminative and generative benchmarks, demonstrating strong performance in both paradigms. 
LEASE consistently outperforms prior unified Self-supervised Learning (SSL) frameworks and achieves competitive or superior results compared to domain-specific and VFM-distillation-based models. 
Our extensive evaluation establishes LEASE as the new SoTA unified SSL framework, introducing dual codebook learning strategies for the efficient unification of discriminative and generative representations. 
Our main contributions are summarized as follows:

\begin{enumerate} % [label={(\arabic*)}, wide=0pt]
% \noindent
% (1) 
\item We introduce \textbf{Learning from Semantic Dictionaries (LEASE)}, the first SSL framework that \textbf{explicitly targets the semantic misalignment between generative and discriminative representations} by jointly learning from paired generative–discriminative codebooks.
% \noindent
% (2) 
\item 
We propose a \textbf{codebook contrast objective} constructed entirely from discriminative centroids. % When c
Combined with token reconstruction, this dual supervision encourages a single encoder to reconcile both semantic paradigms within a unified latent space.
% \noindent
% (3) % Unlike prior VQGAN based unified methods, 
\item 
\textbf{LEASE operates entirely on precomputed discrete tokens}, requiring no online tokenizer, data augmentations, or % and no 
dual encoder architectures, % . This design yields 
yielding \textbf{significantly faster training, reaching 48.7\% and 8.75\% speedup over MAGE and Sorcen respectively.} % the previous unified SoTA.
\item 
Extensive experiments show that \textbf{LEASE outperforms prior unified frameworks (e.g., MAGE, Sorcen), MIM methods and competes with specialized discriminative, generative, and VFM-distillation methods, often surpassing them}. 
LEASE also demonstrates strong robustness, transferability, and competitive conditional generation without additional pretraining.
\end{enumerate}

\section{Related Works}

\begin{figure*}[t]
    \centering
    \includegraphics[width=0.85\textwidth]{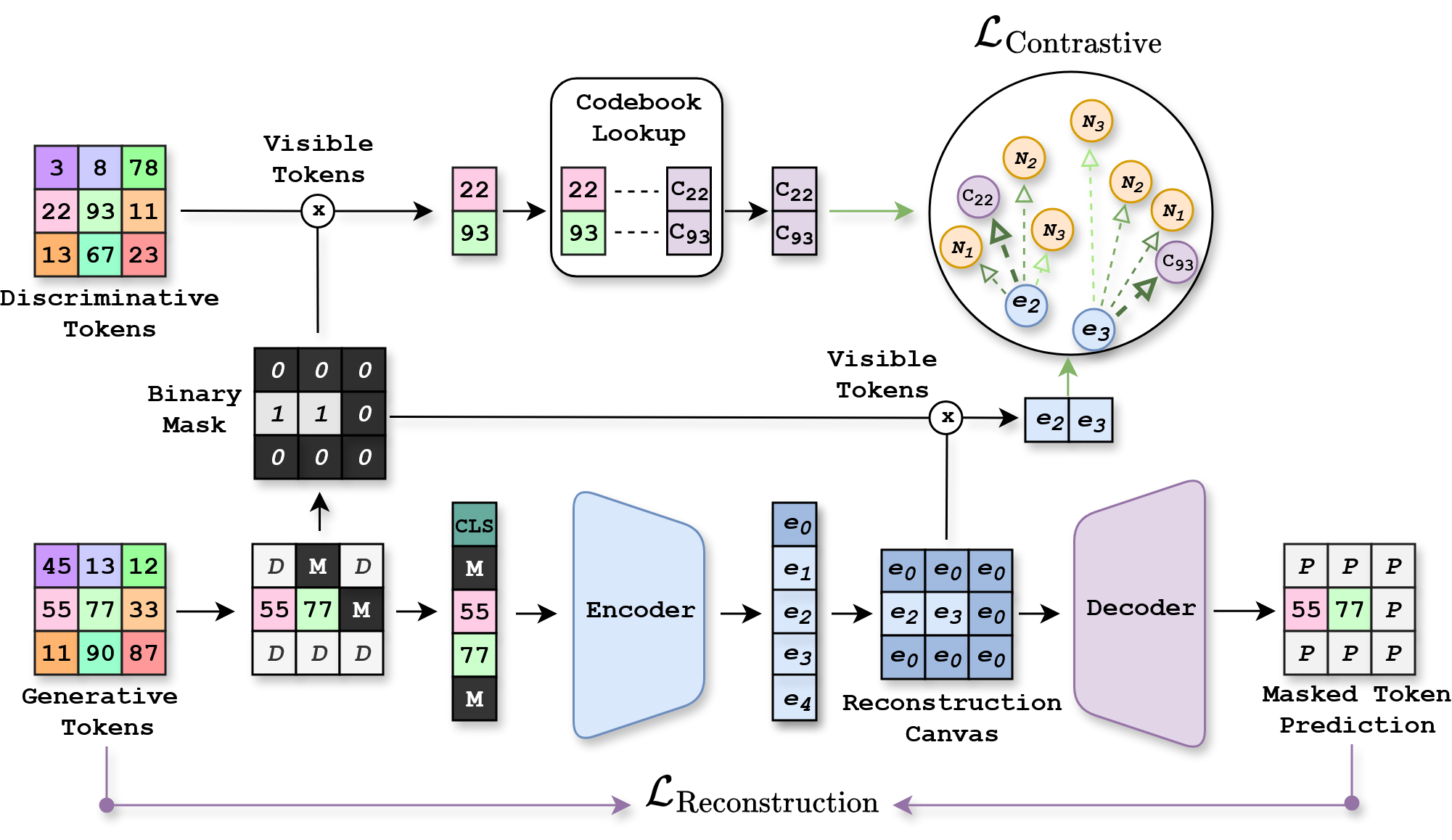}
    \caption{\textbf{Overview of LEASE.} Visible generative tokens are encoded and placed into a reconstruction canvas for masked token prediction, forming \textcolor{Purple}{the reconstruction objective}. Parallelly, visible discriminative tokens are mapped to their codebook centroids, providing positives and neighbors for \textcolor{OliveGreen}{the codebook contrastive loss}, where visible token embeddings and centroids are pulled together. }
    \label{method}
\end{figure*}

\textbf{Self-Supervised Representation Learning.} 
\textbf{Contrastive learning frameworks} have emerged as highly effective unsupervised pretraining methods \cite{caron_deep_2018, grill_bootstrap_2020, chen_simple_2020}. 
SimCLR \cite{chen_simple_2020} established a strong baseline using dual-view augmentations to form positive pairs
%with its dual augmentation pipeline to create positive pairs 
and an InfoNCE-based objective \cite{oord_representation_2019} that attracts these pairs together while repelling % pushing apart 
all remaining representations in the batch. 
MoCo \cite{caron_emerging_2021, chen_empirical_2021} improved this paradigm through momentum encoders and feature queues, although it still suffered from limited sample diversity. 
This motivated the creation of multiple neighbor-based contrastive frameworks \cite{dwibedi_little_2021, koohpayegani_mean_2021, estepa_all4one_2023}. 
NNCLR \cite{dwibedi_little_2021} replaces one member of the positive pair by its nearest neighbor, improving overall semantic diversity. 
% While effective, these contrastive approaches primarily extract the 
While contrastive methods capture strong global semantics,
%, providing general features 
%that excel on simple probes (k-NN or linear probing) but 
%that and
they fall short on complex downstream tasks, such as full finetuning. 
% In contrast, reconstruction-based SSL excels in these tasks. 
\textbf{Reconstruction-based SSL methods} address this gap.
SimMIM \cite{xie_simmim_2022} and MAE \cite{he_masked_2022} introduced masked image modeling (MIM). MAE avoids encoding masked patches, % does not process masked patches by its encoder,
thereby improving training efficiency while obtaining competitive performance.
%results competitive with contrastive learning.
%The use of masked tokens, which are not intially processed by the encoder, enable a more efficient training while obtaining comparable or better results w.r.t contrastive approaches. 
The widespread adoption of Transformer architectures \cite{vaswani_attention_2017} accelerated the development of new MIM architectures. 
U-MAE \cite{zhang_how_2022} and Ge2-AE \cite{liu_devil_2023} address feature collapse and over-smoothing in MAE \cite{he_masked_2022}. 
A2MIM \cite{li_architecture-agnostic_2023} explores model-agnostic MIM strategies that can also be applied to CNNs. 
I-JEPA \cite{assran_self-supervised_2023} and XTRA \cite{amrani_sample-_2024} push state-of-the-art with more predictive and efficient SSL architectures. 
CAE \cite{chen_context_2024}, CMAE \cite{huang_contrastive_2024} and iBot \cite{zhou_ibot_2022} combine MIM, contrastive objectives and distillation, improving representation quality. 
\textit{Different from existing representation learning approaches, LEASE works entirely in a compact, precomputed token space, eliminating the need for %  that is independent of 
augmentations. 
Its codebook contrast objective extracts both positive and negative pairs from a discriminative codebook, eliminating batch-dependent negatives 
% reducing the noise associated with batch sampling 
% reducing 
and the associated sampling noise, while providing required diversity.}

\noindent
\textbf{Unified Visual Representation Learning.} 
While some MIM models exhibit limited generative ability, % capabilities, 
their performance is insufficient for true unified modeling \cite{tian_addp_2024, chen_deconstructing_2024, mukhopadhyay_text-free_2023}. 
Generative models such as Generative Adversarial Networks (GANs) or diffusion-based models produce high-fidelity images in both unconditional and conditional settings. 
Works such as BigGAN \cite{brock_large_2018}, IC-GAN \cite{casanova2021instance} and HiT \cite{zhao_improved_2021} focus on improving generative quality, while DiT \cite{peebles_scalable_2023} introduced the Transformer architecture \cite{vaswani_attention_2017} to diffusion models with strong results. 
l-DAE \cite{chen_deconstructing_2024} explored whether diffusion models can also support discriminative tasks. 
BigBiGAN \cite{donahue_large_2019}, MaskGIT \cite{chang_maskgit_2022} and more recently ADDP \cite{tian_addp_2024}, demonstrate that generation-oriented models can also yield competitive discriminative representations. 
MAGE \cite{li_mage_2023} and Sorcen \cite{estepa_conjuring_2025}, the current unified SSL SoTA, leverage generative tokenizers such as VQGAN \cite{esser_taming_2021} to reconstruct tokenized input images, enhancing both discrimination and generation. 
DiGIT \cite{zhu_stabilize_2024} discretizes features from VFMs and trains next-token prediction models on top, achieving strong unified performance. 
However, most of these approaches require an online tokenizer during training, significantly increasing computational cost. 
Sorcen \cite{estepa_conjuring_2025} partially alleviates this through precomputation but relies on a dual-encoder architecture, which introduces an additional forward pass per training step. Recent methods such as REPA \cite{yu_representation_2024}, VFMTok \cite{zheng_vision_2025}, MergeVQ \cite{li_mergevq_2025} and SVG \cite{shi_latent_2025} use pretrained VFMs as teachers to distill discriminative features into generative frameworks, obtaining SoTA performance. 
While effective, these approaches rely on % require from 
frozen VFMs during training, % which reduces 
reducing the efficiency. 
Although multimodal LLMs can exhibit strong unified capabilities, their scale makes experimentation costly \cite{li_synergen-vl_nodate, ma_unitok_2025}. 
\textit{Our work targets the root cause of unification, the semantic misalignment between generative and discriminative representations. 
LEASE leverages a paired generative–discriminative codebook to train a standalone encoder–decoder model that jointly learns both semantic types. 
Its token reconstruction objective works over precomputed tokens, eliminating the need for online tokenizers and enabling efficient generative training.
% enabling efficient generative training without any online tokenizer. 
Simultaneously, the codebook contrast objective regularizes the model using a discriminative codebook, without requiring dual encoders or distillation from frozen VFMs. 
This design maximizes unified learning while minimizing computational overhead, yielding % leading to 
improvements over prior unified SSL frameworks such as MAGE \cite{li_mage_2023} and Sorcen \cite{estepa_conjuring_2025} while being significantly more efficient.}

\section{Method}
\label{sec:method}

Our Self-supervised Learning Framework, 
\textbf{\ul{LEA}}rning from \textbf{\ul{SE}}mantic Dictionaries (LEASE), 
leverages encoded information from a paired set of generative and discriminative codebooks to train a \emph{unified encoder} capable of both representation learning and visual generation. 
The generative codebook provides input tokens and enables capturing subtle visual details through a \textbf{masked token reconstruction objective}, while the discriminative codebook enriches each patch embedding with strong semantic information via a \textbf{multi-target contrastive learning objective}. Together, these objectives enable the encoder to reconcile generative and discriminative semantics within the same latent space, providing representations that are simultaneously semantically meaningful and rich enough for visual generation and discriminative tasks.
%This duality provides a balanced learning that promotes the domain unification.
%Our novel self-supervised method, \textbf{L}earning from \textbf{S}emantic \textbf{D}ictionaries (LEASE), leverages encoded information of a selected pair of Codebooks, generative and discriminative,  to train a unified encoder that excels in both representation learning and visual generation, thanks to its learned high-level image features that also contain the fine-grained details required for image generation. While the generative Codebook provides the input information and encourages the learning of detailed information via \textbf{Masked Token Reconstruction objective}, the discriminative Codebook ensures that strong semantical information is attached to each patch embedding by the use of a multi-objective \textbf{Contrastive learning based objective}. This duality provides a balanced learning that greatly encourages the domain unification.

\noindent
\textbf{Architecture Overview.}
LEASE (see Figure \ref{method}) consists of a Transformer Encoder $\mathcal{E}$ that projects input semantic tokens into a unified latent space, and a Transformer Decoder $\mathcal{G}$ % capable of transforming 
that transforms the unified latent representations back into semantic tokens. 
%Different to 
Unlike popular MIM architectures, LEASE % is designed to work with
operates directly on quantized tokens instead of raw image pixels, making them computationally cheaper and enabling %substantially 
faster pretraining. 
This % training 
efficiency is further enhanced through codebook-based dataset precomputation, where the entire dataset is tokenized \emph{just once} before training, eliminating any on-the-fly tokenization overhead.
%and remove in this way any possible training time overhead caused by this tokenization process.

\noindent
\textbf{Codebooks.}
LEASE leverages two complementary codebooks during training: a generative codebook and a discriminative codebook. 
The generative codebook acts as the tokenizer, transforming the original input image into a discrete sequence of \textit{N} tokens, each token represented by integer indices. 
Previously trained for image reconstruction, it encodes the image into detailed semantics useful for generation. 
On the other hand, the discriminative codebook, which is derived from a self-supervised VFM, provides strong discriminative semantics for the model to learn and align with input generative tokens.
%In contrast to the generative codebook, the discriminative codebook captures powerful and dense discriminative information.
Rather than being trained, this codebook is created by clustering the embedding space into $K$ centroids, each representing a distinct semantic concept. 
For every image patch, the two codebooks produce complementary information that together form the foundation of LEASE’s unified modeling framework.

%\textbf{Codebooks. }LEASE leverages two different types of codebooks during training: generative and discriminative. Generative codebook acts as a tokenizer, transforming the original input image into a discrete sequence of \textit{N} tokens or semantics, represented by integers. Trained for image reconstruction, this generative codebook summarizes the image into detailed semantics useful for generation. Discriminative codebook, in contrast, contains powerful and dense discriminative information. Instead of being trained, this codebook is directly extracted from an SSL Vision Foundation Model by clustering its embedding space into \textit{K} different semantics, which are represented by the \textit{K} clustering centroids. For every patch in an image, these codebooks let us extract two streams of information that, while different, provide the base for a truly unified model.

\noindent
\textbf{Input Preparation.}
We precompute the discrete token sequences of the target dataset $\mathcal{D}$ using both the generative and discriminative codebooks. 
For each image, $\mathcal{I}$, we encode every image patch into a sequence of generative semantic tokens, $t = (t_1, \ldots, t_{SS})$, where $SS$ denotes the sequence length (number of patches). 
Each value in $t$ is represented by an integer in the range $[0, v_{max}]$, 
%being $v_{max}$ 
where $v_{max}$ denotes the vocabulary size of the generative tokenizer. 
In parallel, we also extract a sequence of discriminative semantic tokens, $t'$, 
%replicating
using the discriminative codebook in an identical fashion.
After preprocessing, each image is associated with 
a pair of token sequences. 
While $t$ serves as the input to LEASE, $t'$ links each generative token to its corresponding discriminative semantic, forming a position-aligned positive pair. Concretely, the $i$-th element of $t$ and the $i$-th element of $t'$ correspond to the same image patch, allowing a direct lookup in the discriminative codebook to retrieve the discriminative semantic associated with $t'_i$. This establishes a simple but effective correspondence chain that provides the required positive pairs for the codebook contrast objective. To maintain a completely unsupervised learning scheme, we use an \enquote{unsupervised} VQGAN as our generative codebook following previous works \cite{li_mage_2023, estepa_conjuring_2025}.

\subsection{Generative Objective}

The first objective enhances LEASE's generative capacity by leveraging the generative semantic tokens through two steps: the \textit{Masking step} and the \textit{Reconstruction step}.

\noindent
\textbf{Masking Step.} 
While high masking ratios benefit generative modelling, lower ratios increase the representation learning capacity of the model \cite{li_mage_2023}. 
LEASE leverages a variable masking ratio strategy that 
%enable both low and high masked sequences
enables learning from both low and high masking ratios, balancing these two objectives. On average, 69\% of the input sequence is masked, ranging between 50\% and 100\%. At each training step, masked tokens are replaced with an integer outside the vocabulary range $v_{max}$, 
%we define 
denoted as mask token \textit{[MASK]}. 
Then, a special token \textit{[CLS]} is prepended to the sequence. Finally, to reduce memory footprint, LEASE drops masked tokens retaining only half of the original sequence size. 
This dropping strategy follows prior works \cite{he_masked_2022, li_mage_2023}. 
After this drop, $\approx$ 19\% of the final sequence size is kept masked. 

\noindent
\textbf{Reconstruction Step.} 
The reconstruction process is formulated as a token prediction problem applied to each masked patch in the input sequence $t_{masked}$. 
First, the LEASE encoder $\mathcal{E}$ projects the $t_{masked}$ into its unified latent space, obtaining  $\mathcal{E}(t_{masked}) = (e_0, e_1,\ldots, e_L)$, where $L$ is the length of $t_{masked}$ and half of the original input sequence length. 
To reconstruct the original input sequence $t$, we create a template "canvas", $cv$, of size $2 \cdot L$ using the [CLS] latent $e_0$. 
Then, the corresponding positions in the canvas are replaced with their latent embeddings from $t_{masked}$.
%we replace positions that represent the unmasked tokens with their latent in $t_{masked}$.
Finally, the canvas is passed through the LEASE decoder $\mathcal{G}$ to predict the original input sequence $t$. 

The token reconstruction objective is defined as:
\begin{equation}
\mathcal{L}_{\text{R}} = -\mathbb{E}_{t \sim \mathcal{D}} 
\left( 
\sum_{i=1}^{CS} m_i \log p(t_i \mid cv_i) 
\right).
\end{equation}
where $p$ represents the output of the decoder $\mathcal{G}$ and $CS$ the canvas size. Note that this objective is exclusively applied to tokens represented as masked, $m_i$, in the canvas. 

\subsection{Discriminative Objective}
This objective leverages the discriminative codebook to enhance the semantic understanding capacity of LEASE and consists of two stages: \textit{centroid gathering} and \textit{codebook contrast}.

\noindent
\textbf{Centroid Gathering.} 
For each token in the input sequence $t$, there exists a corresponding discriminative token in sequence $t'$. 
This alignment correspondence enables an efficient lookup over the selected discriminative codebook to obtain its centroid $\mathcal{C}_{t'_i}$, represented as a semantic latent. In practice, for each latent $e_i$ in LEASE space representing a token $t_i$, the centroid $\mathcal{C}_{t'_i}$ is obtained through the mapping $e_i\rightarrow t_i\rightarrow t'_i \rightarrow \mathcal{C}_{t'_i}$. 
To enrich semantic information, $K_{sel}$ most similar centroids to $\mathcal{C}_{t'_i}$ are retrieved from the entire set of $K$ centroids, forming the set of neighbor centroids $\mathcal{N}_{t'_i}$. 
The retrieval is 
%efficiently 
computed for all tokens 
%in the batch 
as follows:
\begin{equation}
 \mathcal{N}_i = \operatorname{TopK}\!\big(\{ s_{ik} \}_{k=1}^{K},\; K_{\mathrm{sel}}\big).   
\end{equation}
where $sim_{ik}$ denotes the cosine similarity between centroids, 
%is denotes
%d by the Centroid-Centroid cosine similarity 
$s_{ik} = {\mathbf{C}}_{t_i}^{\top}{\mathbf{C}}_{k}$. 
Each token in the batch thus obtains $1+\mathcal{N}_{t'_i}$ positive pairs, enabling our novel codebook based contrastive objective.

\noindent
\textbf{Codebook Contrast.}
While the centroid $\mathcal{C}_{t'_i}$ and its neighbors $\mathcal{N}_{t'_i}$ are considered positive samples, the semantic similarity between the original $\mathcal{C}_{t'_i}$ and its neighbors may vary significantly. 
For this reason, our codebook contrast objective adaptively weighs the relevance of each neighbor based on their similarity to the query centroid $\mathcal{C}_{t'_i}$. 
For every neighbor $j$ in $\mathcal{N}_{t'_i}$ we obtain a weight based on the previously computed similarity:
\begin{equation}
w_{ij} = 
\frac{
    \exp\left( sim_{ij}/\tau
    \right)
}{
    \sum\limits_{k \in\mathcal{C}_{t'_i} \cup \mathcal{N}_{t'_i}} 
    \exp\left(sim_{ik}/\tau
    \right)
}
\end{equation}
where $\tau$ controls the smoothness of the weighting distribution. 
It is set to 0.1 to keep the target sharp while preserving meaningful weights for close centroids.

Given these weights, the final Codebook Contrast is computed over the unmasked tokens:
\begin{equation}
\mathcal{L}_{\text{C}} 
= 
- \frac{1}{N_u} 
\sum_{i \in \mathcal{U}}
\sum_{j \in \mathcal{C}_{t'_i} \cup \mathcal{N}_{t'_i}}
w_{ij}\,
\log
\frac{
    \exp\!\left( z_i^{\top}{\mathcal{C}}_{j} / \alpha
    \right)
}{
    \sum\limits_{k=1}^{K}
    \exp\!\left(
        z_i^{\top}{\mathcal{C}}_{k} / \alpha
    \right)
}
\end{equation}
where $N_u$ is the number of unmasked tokens and $\alpha$ is the contrastive temperature. 
This objective pulls together the unified semantics produced by the LEASE encoder and those encoded in the discriminative codebook, encouraging the model to produce highly semantic embeddings. 
Unlike standard contrastive objectives, Codebook Contrast does not rely on batch samples to form negative pairs. Instead, it uses all remaining centroids in the codebook, those not selected as neighbors or as $\mathcal{C}_{t'_i}$, as negatives. This removes the noise and instability of batch-dependent negatives, yielding more stable and meaningful representations.

\subsection{Final Objective}
The overall LEASE objective is defined as the weighted sum of its Reconstruction and Contrast losses:
\begin{equation}
L_{LEASE} = L_{R} + \lambda \cdot L_{C},
\end{equation}
where $\lambda$ controls the contribution of the discriminative signal during training. 
This unified objective effectively integrates representation learning and visual generation, resulting in a model that excels in both paradigms, ultimately demonstrating strong SoTA results across both.

\section{Experiments} % [WIP: I need to tweak some analysis]}
\label{sec:experiments}

\textbf{Implementation Details.}
All experiments use ViT-Base architecture following the configuration reported in MAGE \cite{li_mage_2023}. 
The generative codebook is obtained from an unsupervised VQGAN \cite{esser_taming_2021}, 
% trained in an unsupervised manner, 
while the discriminative codebook is constructed by applying k-means clustering to DINOv2 \cite{oquab_dinov2_2024} features, following the procedure introduced in DiGIT \cite{zhu_stabilize_2024}. 
Unless otherwise stated, all models are pretrained on IN-1K \cite{deng_imagenet_2009} for 1600 epochs and evaluated on its validation set.  
For unconditional generation, we adopt the evaluation protocols used in MAGE \cite{li_mage_2023} and MaskGIT \cite{chang_maskgit_2022}. 
Additional implementation details and % , including 
hyperparameters are provided in the supplementary material.

\subsection{Unified Evaluation}
% We compare LEASE against state-of-the-art generative, discriminative, and unified SSL frameworks on ImageNet-1K \cite{deng_imagenet_2009} across linear probing and unconditional image generation. 

We evaluate LEASE on IN-1K \cite{deng_imagenet_2009} using linear probing and unconditional image generation, and compare it against SoTA generative, discriminative, and unified SSL methods (See \Cref{megatable}). 
% Results are presented in Table 1.
% Results are summarized in Table \ref{megatable}. 
When compared against \textit{\textbf{generative methods}}, LEASE achieves competitive or superior performance in both FID and IS. 
It outperforms all generative baselines except ADDP \cite{tian_addp_2024}, while LEASE substantially outperforms it in discriminative accuracy. This highlights that models optimized primarily for generation struggle to match LEASE’s balanced unified capability. 
Compared to SSL \textit{\textbf{contrastive methods}},
%known for their representation learning capacity, 
LEASE ranks among the top. 
Only MAGE-C achieves higher linear probing accuracy, but LEASE outperforms it in generation quality. 
MAGE-C introduces a contrastive objective that prioritizes discriminative performance at the expense of generation, reducing its unified effectiveness. 
LEASE consistently outperforms all \textit{\textbf{MIM methods}} while being unified. 
Among VQGAN-based unified models, LEASE performs on par with Sorcen \cite{estepa_conjuring_2025} and MAGE \cite{li_mage_2023} in unconditional generation. However, LEASE’s latent space provides stronger discriminative features and a higher linear-probe accuracy. 
Sorcen %, in particular, 
matches LEASE in generation but is consistently outperformed on discriminative evaluation. Overall, LEASE displays solid performance in both discriminative and generative tasks, outperforming prior unified VQGAN-based models and MIMs while remaining competitive with specialized single-domain methods.

\begin{table}[t!]
\small
\centering
\renewcommand{\arraystretch}{0.9}
\begin{tabular}{llllccc}
\toprule
Method        & Models & LP\%                                                & FID                                                 & IS    \\ \midrule
\multicolumn{5}{l}{\hphantom{...........................} \textit{(Generative models)}}                                                                                                      \\
BigGAN        &        & -                                                   & 38.6                                                & 24.70 \\
BigGAN+ Clust &        & -                                                   & 22.0                                                & 23.50  \\
HiT           &        & -                                                   & 30.8                                                & 21.64 \\
ADM           &        & -                                                   & 26.2                                                & 39.70 \\
MaskGIT       &        & 57.4                                                & 20.7                                                & 42.08 \\
BigBiGAN      &        & 56.6                                                & 21.6                                                & 27.94 \\
IC-GAN        &        & -                                                   & 15.6                                                & 59.00  \\
ADDP          &        & 11.5                                                & 8.9                                                 & 95.32 \\
DiT           &        & 62.5                                                & 30.9                                                & -     \\
l-DAE         &        & 69.6                                                & -                                                   & -     \\ \midrule
\multicolumn{5}{l}{\hphantom{.............................}\textit{(Contrastive Models)}}                                                                                                     \\
SimCLRv2      & R50-w2 & 75.6                                                & -                                                   & -     \\
MoCov3        & ViT-B  & 76.7                                                & -                                                   & -     \\
NNCLR         & ViT-B  & 76.5                                                & -                                                   & -     \\
DINO          & ViT-B  & 72.8                                                & -                                                   & -     \\
iBot          & ViT-B  & 76.0                                                & -                                                   & -     \\
MAGE-C        & ViT-B & 78.2                        & 31.8                                                & 37.40 \\
CMAE          & ViT-B  & 73.9                                                & -                                                   & -     \\ \midrule \midrule
\multicolumn{5}{l}{\hphantom{...................................}\textit{(MIM models)}}                                                                                                             \\
MAE           & ViT-B  & 68.0                                                & -                                                   & -     \\
SimMIM        & ViT-B  & 57.8                                                & -                                                   & -     \\
U-MAE         & ViT-L  & 65.8                                                & -                                                   & -     \\
BeiT          & ViT-B  & 56.7                                                & -                                                   & -     \\
CAE           & ViT-B  & 70.4                                                & -                                                   & -     \\
%XTRA          & ViT-B  & 70.2                                                & -                                                   & -     \\
I-JEPA        & ViT-B  & 72.9                                                & -                                                   & -     \\
Ge2-AE        & ViT-B  & 75.3                                                & -                                                   & -     \\
A2MIM         & ViT-B  & 68.8                                                & -                                                   & -     \\
XTRA          & ViT-B  & 70.2                                                & -                                                   & -     \\ \midrule
Sorcen        & ViT-B  & 75.1                                                & \textbf{9.61}                                                & 90.96 \\
MAGE          & ViT-B  & 74.7                                                & 11.1                                                & 81.17 \\

MAGE$\dagger$    & ViT-B  & 75.0                                                & 10.88                                               & 81.59 \\ \midrule
\rowcolor{sorcenrow} LEASE (Ours)          & ViT-B  & \textbf{76.7} & \textbf{9.62} & \textbf{91.78} \\ \bottomrule
\end{tabular}
\caption{Unified evaluation % results reported on 
using linear probing (LP\%) and unconditional generation (FID/IS). $\dagger$ - reproduced results.}\label{megatable}
\end{table}

\subsection{VQGAN-based Model Comparison}
We further compare % analyse the performance of LEASE against 
LEASE with VQGAN \cite{esser_taming_2021} based unified methods, MAGE \cite{li_mage_2023} and Sorcen \cite{estepa_conjuring_2025} across four axes:
% This evaluation is conducted along four axes: 
(1) performance in low-data regimes, % scenarios, 
(2) full finetuning task on IN-1K, (3) robustness to out-of-distribution and corrupted data, and (4) transferability across datasets. 

\noindent
\textbf{Low-data Regimes (\Cref{fewshot}).}
We linear probe every model for 10 epochs (5, 10, 13 \& 25 shots per class), while keeping the rest of hyperparameters fixed. 
% As shown in Table \ref{fewshot}, 
LEASE demonstrates superior performance in few-shot settings, achieving an average improvement of 2.32\% over MAGE and 0.56\% over Sorcen. 
The dual nature of its training objectives provides a richer and more informative latent space, enabling the model to make better use of limited labeled samples and improving its effectiveness in low-data regimes.

% Few-shot
% Transfer Learning
% Robustness
% Robustness

\begin{table}[t!]
\small
\centering
\renewcommand{\arraystretch}{0.7}
\begin{tabular}{@{}lccccc@{}}
\toprule
\multicolumn{1}{l}{\multirow{2}{*}{Method}} & \multicolumn{5}{c}{Shots per IN-1K Class}                                 \\ \cmidrule(l){2-6} 
\multicolumn{1}{l}{}                        & 5              & 10             & 13             & 25             & Avg. \\ \midrule
MAGE                                        & 48.44          & 57.37          & 59.84          & 63.66          & 57.33    \\
Sorcen                                      & 50.30          & 59.21          & 61.73          & 65.13          & 59.09    \\
LEASE                                        & \textbf{51.00} & \textbf{59.39} & \textbf{62.00} & \textbf{66.19} & \textbf{59.65}    \\ \bottomrule
\end{tabular}
\caption{Top-1 Accuracy (\%) for few-shot on IN-1K. % and average.
} \label{fewshot}
\end{table}

\noindent
\textbf{Fine tuning Performance (\Cref{finetune}).}
We finetune LEASE on IN-1K following % the setup reported by 
MAGE \cite{li_mage_2023}. 
% As reported in Table \ref{finetune}, 
All three frameworks achieve comparable results under full fine-tuning. 
Nevertheless, LEASE slightly outperforms the others, achieving a 0.2\% gain over MAGE \cite{li_mage_2023}, the strongest baseline. 

\begin{table}[!t]
\small
\centering
\renewcommand{\arraystretch}{0.7}
\begin{tabular}{@{}lcc@{}}
\toprule
Method        & Top-1 & Top-5 \\ \midrule
MAGE  & 82.5      & -         \\
Sorcen        & 82.4      & 96.0         \\
LEASE         & \textbf{82.7}      & \textbf{96.1}         \\ \bottomrule
\end{tabular}
\caption{Accuracy (\%) for fine-tuning on IN-1K.} \label{finetune}
\end{table}

\noindent
\textbf{Transferability (\Cref{tab:transfer16}).} % of the model.}
We evaluate cross-dataset transferability on Caltech-101 \cite{fei2007learning}, UCF-101 \cite{soomro2012ucf101}, Sun397 \cite{xiao2010sun}, DTD \cite{cimpoi2014describing}, CIFAR-10 \cite{krizhevsky2009learning}, CIFAR-100 \cite{krizhevsky2009learning}, and Places365 \cite{zhou2017places}. 
% As shown in Table \ref{tab:transfer16}, 
LEASE outperforms other baselines on 5/7 datasets, yielding an average improvement of 0.75\% over MAGE \cite{li_mage_2023} and 0.89\% over Sorcen \cite{estepa_conjuring_2025}. 
These results indicate that LEASE learns more generalizable representations that transfer effectively beyond the pretraining domain.

\begin{table}[t!]
\small
\centering
\setlength{\tabcolsep}{2pt}
\begin{tabular}{@{}lccccccccc@{}}
\toprule
              & Caltech & UCF101 & Sun   & DTD   & C100 & C10 & Places & Avg.        \\ \midrule
MAGE          & 88.97   & 59.66  & 52.36 & \textbf{53.90}  & 61.39        & \textbf{83.60}       & 32.35         & 61.75 \\
Sorcen        & 89.61   & \textbf{62.44}  & 53.26 & 53.84 & 59.00    & 80.27   & 32.82     & 61.61 \\
LEASE         & \textbf{89.82}   & 62.23  & \textbf{54.71} & 50.53 & \textbf{63.61}    & 81.21   & \textbf{35.36}     & \textbf{62.50}       \\ \bottomrule
\end{tabular}
\caption{Transfer learning results (Top-1 accuracy (\%)) for different datasets under 16-shot settings. % and average across datasets.
}
\label{tab:transfer16}
\end{table}

\noindent
\textbf{Model Robustness (\Cref{tab:robustness}).}
We assess model robustness by evaluating the learned representations using % a 
k-NN % probe 
on six robustness benchmarks % containing challenging, distorted, out-of-distribution, and corrupted samples 
\cite{recht_imagenet_2021, wang_learning_2019, hendrycks_many_2021, hendrycks_natural_2021, barbu_objectnet_2019, hendrycks_benchmarking_2018}. 
For ImageNet-C \cite{hendrycks_benchmarking_2018}, we report results at corruption severities 4 and 5, which represent the most difficult settings. 
Additional dataset details are provided in the supplementary material. 
For reference, we also include the k-NN performance on the IN-1K validation set. 
Across all robustness evaluations, LEASE clearly outperforms the baselines, demonstrating its ability to produce more stable and resilient features.

\begin{table}[t!]
\small
\centering
\setlength{\tabcolsep}{2pt}
\begin{tabular}{@{}lcccccccc@{}}
\toprule
              & Val. & v2 & IN-S & IN-R  & IN-A & ObjN. & S4 & S5 \\ \midrule
MAGE & 60.65    & 45.94   & 14.09     & 22.32 & 4.19 & 10.31     & 21.96         & 13.69         \\
Sorcen        & 62.23    & 48.35   & 14.62     & 25.27 & 5.47 & 13.52     & 22.15         & 14.42         \\
LEASE          &   \textbf{62.60}       &     \textbf{49.14}    &      \textbf{19.48}     &    \textbf{28.04}   &    \textbf{8.51}  &  \textbf{21.61}         & \textbf{30.24}         & \textbf{20.39}         \\ \bottomrule
\end{tabular}
\caption{K-NN Top-1 accuracy (\%) in IN-1K robustness variants.}
\label{tab:robustness}
\end{table}

\begin{figure}[t]
    \centering
    \includegraphics[width=0.45\textwidth]{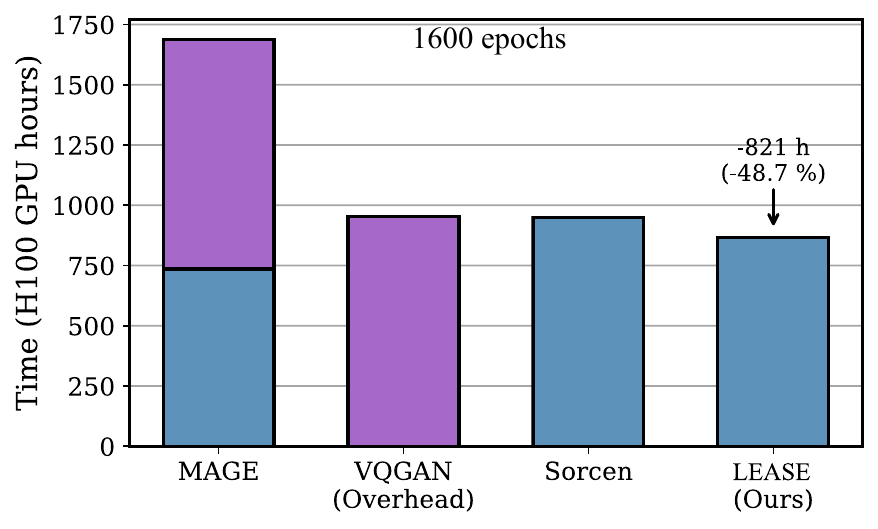}
    \caption{\textbf{Training time} for VQGAN-based unified methods. % comparison for 1600 epoch of VQGAN-based unified methods
    }
    \label{fig:timebar}
\end{figure}

\noindent
\textbf{Framework Efficiency (\Cref{fig:timebar}).} 
% In Figure \ref{fig:timebar} 
We evaluate the training time of each unified model under an identical setup using a single H100 GPU, a batch size of 128, IN-1K \cite{deng_imagenet_2009} for pretraining and FlashAttentionv2 \cite{dao_flashattention-2_2023} for optimized operations. 
MAGE \cite{li_mage_2023} relies on an online tokenizer (highlighted in purple), which significantly increases its computational overhead and leads to substantially longer training times. 
Sorcen \cite{estepa_conjuring_2025} removes the need for online tokenization by precomputing its inputs but still requires an additional forward pass to compute its contrastive objective, resulting in slower training compared to a single-pass design. 
In contrast, LEASE requires only one forward pass per iteration, and both of its objectives operate solely on lightweight codebooks, obtaining a more efficient training process. % that MAGE \cite{li_mage_2023} and even Sorcen \cite{estepa_conjuring_2025}, who also precomputes its inputs. 
Under this shared setup, LEASE trains 48.7\% faster than MAGE \cite{li_mage_2023} and 8.75\% faster than Sorcen \cite{estepa_conjuring_2025}.

\subsection{DINOv2-based Model Comparison}
% Given that 
Since LEASE leverages a discriminative codebook derived from DINOv2 \cite{oquab_dinov2_2024}, we further compare (\Cref{dinov2_comparison}) it 
% in Table \ref{dinov2_comparison} 
to models that incorporate DINOv2 knowledge through different learning strategies. 
We evaluate all methods on both unconditional generation (FID/IS) and discriminative performance (linear probing and fine-tuning). 
Among approaches that directly distill features from a DINOv2 encoder, LEASE’s codebook contrast provides a stronger alternative, outperforming REPA \cite{yu_representation_2024}, SVG \cite{shi_latent_2025}, and VFMTok \cite{zheng_vision_2025} across both discriminative and generative metrics. MergeVQ \cite{li_mergevq_2025}, in its unified (G+R) configuration that also distills from DINOv2, achieves higher linear probing accuracy but underperforms in fine-tuning. DiGIT \cite{zhu_stabilize_2024}, which also uses a discriminative codebook to construct input sequences, lags behind LEASE in linear probing and IS. Although DiGIT reports a lower FID, LEASE’s codebook contrast objective offers a superior balance of performance and efficiency, as it does not need a VFM during training.

\begin{table}[t!]
\centering
\small
\setlength{\tabcolsep}{3pt}
\begin{tabular}{@{}llcccc@{}}
\toprule
Method       & Learning type     & FID  & IS    & LP\% & FT\%  \\ \midrule
DINOv2       & ViT-B             & -    & -     & 84.5 & 85.7  \\ \midrule
REPA         & Distillation      & -    & -     & 68.2 & -     \\
DiGIT        & Codebook Input    & 9.13 & 73.85 & 71.7 & -     \\
SVG          & Init/Distill      & -    & -     & -    & 81.8 \\
VFMTok       & Distillation      & -    & -     & 69.4 & -     \\
MergeVQ (G+R) & Distillation      & -    & -     & 77.9 & 82.0  \\
\rowcolor{sorcenrow} LEASE (Ours)         & Codebook Contrast & 9.62 & 91.78 & 76.7 & 82.7  \\ \bottomrule
\end{tabular}
\caption{Unified evaluation against frameworks based on DINOv2.}\label{dinov2_comparison}
\end{table}

\subsection{Conditional Generation}
% While LEASE has been mainly designed for complete unsupervised pretraining, its knowledge can be extended to conditional generation. 
Although LEASE is primarily designed for unsupervised pretraining, it can be adapted for conditional generation.
Using the same architecture, we finetune the decoder 
% is finetuned 
using an additional class embedding token extracted from CLIP \cite{radford_learning_2021} and a sequence tail formed by all centroids extracted from the discriminative codebook. 
This finetuning phase is optimized by the reconstruction loss exclusively, as the encoder is kept completely frozen. 
More setup details can be found in the supplement. 
Despite this constrained % conditional generation 
setup, LEASE achieves competitive conditional generation performance (\Cref{tab:conditional}).
% performs competitively with state-of-the-art methods. 
Notably, it outperforms MAGE \cite{li_mage_2023} on FID while keeping the IS competitive without any from-scratch pretraining and a smaller decoder. It also performs better than DiGIT \cite{zhu_stabilize_2024} and REPA \cite{yu_representation_2024}, two models that also leverage DINOv2 during training. 
While MergeVQ \cite{li_mergevq_2025} reports the strongest results,
% Although MergeVQ \cite{li_mergevq_2025} achieves the strongest overall performance, 
LEASE attains a similar % closely matches its 
FID with fewer parameters. 
% while using fewer model parameters. 
% This analysis further shows the capacity of out encoder to produce representations that provide meaningful information even if has been acquired exclusively from an unsupervised training schedule.
These findings indicate that LEASE’s encoder produces strong semantic representations even when trained entirely without supervision.

\begin{table}[t!]
\centering
\small
\setlength{\tabcolsep}{3pt}
\begin{tabular}{@{}llllcc@{}}
\toprule
Type  & Tokenizer    & Generator            & Parameters & FID  & IS     \\ \midrule
Mask. & VQGAN        & MaskGIT              & 177M       & 6.18 & 182.1  \\
Mask. & VQGAN        & MAGE                 & 117M+113M  & 6.93 & 195.8  \\
AR    & VQGAN        & DiGIT                & 219M       & 4.79 & 142.87 \\
Diff. & -            & REPA (SiT-L/2)       & 458M       & 9.90 & 111.9  \\
AR    & MergeVQ & MergeAR              & 343M       & 3.25 & 253.8  \\
\rowcolor{sorcenrow} Mask. & VQGAN        & LEASE                 & 117M+79M   & 3.72 & 179.09 \\\bottomrule
\end{tabular}
\caption{Results on Conditional Generation (FID and IS).}
\label{tab:conditional}
\end{table}

\subsection{Ablations and Codebook Analysis}

\noindent
\textbf{General Ablations (\Cref{main_ablation}).} 
% We ablate the components of LEASE in Table \ref{main_ablation}. 
% When comparing against a 
Compared to pure reconstruction baseline without the codebook contrast objective, we observe consistent improvements across all metrics, demonstrating that the proposed contrastive objective enhances both generative and discriminative performance. 
% This confirms that aligning semantics in a unified latent space benefits both tasks. 
We further analyze \textit{where} the contrastive supervision should be applied. 
Applying the codebook contrast on the decoder features improves discriminative accuracy but hinders generative quality (second row of \Cref{main_ablation}), indicating that unification must occur within the encoder’s latent space. 
Finally, combining contrastive supervision on both encoder and decoder does not provide additional discriminative gains beyond encoder-only contrast, while slightly degrading generation performance. 
% Given these observations, 
Thus LEASE is conformed by the reconstruction loss and the codebook contrast on the encoder's feature space. 
% More hyperparameter ablations can be found in supplement.
Additional hyperparameter ablations are provided in the supplement.

%We ablate LEASE in Table xxxx. When compared against pure reconstruction approach, without the Codebook contrast objective, we can see how this novel objective improves all metrics while being constructed over a discriminative codebook. This proves how a unified space can improve both discrimination and generation capacity of the model. Still, we show how this unification process should be done on the space provided by the encoder. If applied over the decoder features, we improve discrimination with the cost of decreasing generation, as it is shown in the second row of Table xxxx. Finally, it is also displayed how the combination of both does not provide any further discriminative improvements from the original Codebook Contrast while being marginally worse on generation. Given these analysis and for the shake of simplicity, LEASE is conformed by the reconstruction loss and the Codebook Contrast on the encoders feature space.

\begin{comment}
\begin{table}[h!]
\centering
\small
\begin{tabular}{@{}llll@{}}
\toprule
                                    & Acc.    & FID   & IS    \\ \midrule
Only Reconstruction          & 73.62 & 10.62 & 79.59 \\
   +DecoderContrast                  & 74.20  & 10.97 & 78.73 \\
PR+CodebookContr.+DecoderContr. & 76.07 & 10.36 & 84.33 \\
LEASE                                & 76.11 & 10.35 & 86.71 \\ \bottomrule
\end{tabular}
\caption{Ablation results reported on linear probing accuracy, FID and IS.} \label{main_ablation}
\end{table}
\end{comment}

\begin{table}[t!]
\centering
\small
\begin{tabular}{llccccccc@{}}
\toprule
                                    & Rec. & DC & EC & LP\%  & FID   & IS    \\ \midrule
                                    & \checkmark &        &        & 73.62 & 10.62 & 79.59 \\
                                    & \checkmark & \checkmark &        & 74.20 & 10.97 & 78.73 \\
                                    & \checkmark & \checkmark & \checkmark & 76.07 & 10.36 & 84.33 \\
\rowcolor{sorcenrow} LEASE                               & \checkmark &        & \checkmark & 76.11 & 10.35 & 86.71 \\ \bottomrule
\end{tabular}
\caption{Ablation results reported on linear probing accuracy, FID and IS. Rec. = Reconstruction Objective, DC = Decoder Contrast, EC = Encoder Contrast.} \label{main_ablation}
\end{table}

\noindent
\textbf{Codebook Size (\Cref{tab:combined} (left)).} 
% We test an additional codebook size on Table \ref{tab:combined} (left). 
Given the % total 
number of 
images on IN-1K, reducing the number of available semantics by half translates into a decreased performance for both discrimination and generation. More detailed patch semantics benefit LEASE by providing a more fine-grained contrast.  

\noindent
\textbf{Codebook Origin (\Cref{tab:combined} (right)).} 
% In Table \ref{tab:combined} (right), 
We analyze the behavior of LEASE using different discriminative codebooks. 
All codebooks are created by extracting patch-level features from the ImageNet-200 subset and clustering them using k-means with $k = 16{,}000$. 
For consistency, all variants are pretrained on IN200 for 200 epochs. 
% Results in Table \ref{tab:combined} (right) correspond to evaluation on IN200. 
As a reference, the first row reports the results obtained with the main DINOv2 codebook obtained from IN-1K. 
Overall, DINOv2 \cite{oquab_dinov2_2024} provides the strongest codebook among all tested VFMs, with the IN200-based variant even outperforming its counterpart created from IN-1K. 
DINOv3 \cite{simeoni_dinov3_2025} ranks second, followed by CLIP \cite{radford_learning_2021} and FRANCA \cite{venkataramanan_franca_2025}.

\begin{table}[t!]
\centering
\small
\renewcommand{\arraystretch}{0.7}
\addtolength{\tabcolsep}{-0.4em}
\begin{tabular}{@{}lccc@{}}
\toprule
 & \multicolumn{2}{c}{Codebook Size} \\
\cmidrule(lr){2-3}
 & 8K & 16K \\ \midrule
LP & 75.20 & 76.70 \\
FID & 10.13 & 9.62 \\
IS & 85.87 & 91.78 \\ \bottomrule
\end{tabular}
\hfill
\begin{tabular}{@{}llll@{}}
\toprule
Origin & LP\%.   & FID  & IS    \\ \midrule
DINOv2$\dagger$ &   85.47   &   16.33   &   58.11    \\ \midrule
FRANCA &   78.91   &    25.36  &   40.88    \\
CLIP   &   79.17   &   25.30   &    41.21 \\
DINOv2 & \textbf{85.86} & \textbf{16.15} & \textbf{59.28} \\ 
DINOv3 & 79.98 & 24.07 & 43.34 \\ \bottomrule
\end{tabular}
\caption{(Left) Comparison between different size codebooks, evaluated in IN-1K. (Right) Comparison between multiple codebook origins. $\dagger$ indicates generated from IN-1K. Otherwise, IN200 was used. Results shown in IN200.}
\label{tab:combined}
\addtolength{\tabcolsep}{+0.4em}
\end{table}
\subsection{Limitations and Future Lines}

% While LEASE demonstrates strong unified performance, several limitations remain to be addressed in future research. First, the current framework is implemented for fully unsupervised training scenarios. Incorporating weak supervision, such as text guidance from a Vision-Language model, could further improve its results in tasks like conditional generation; however, this would compromise the self-supervised nature of the model. Second, the hyperparameters for the codebook contrast, such as the number of neighbors and the temperature parameter ($\tau$), have been selected for general datasets. Domain-specific datasets may require additional parameter tuning. Finally, even if the token precomputation step drastically reduces training time, it does not eliminate the overhead of online tokenization during downstream tasks. Future work may explore architectural modifications to bypass the tokenizer entirely, further improving the inference efficiency.

While LEASE achieves strong unified performance, several limitations remain. 
The current framework is fully unsupervised; incorporating weak supervision (text guidance) could improve conditional generation but would compromise the self-supervised setting. 
The codebook-contrast hyperparameters (number of neighbors, temperature $\tau$) are tuned for general datasets and may require adjustment for specific domains. 
% Finally, 
Although token precomputation reduces training cost, downstream tasks still require online tokenization. 
Future work may explore architectures that bypass the tokenizer entirely to further improve inference efficiency.

\section{Conclusions}

% This work 
We presented LEArning from SEmantic Dictionaries (LEASE), a unified SSL framework that addresses the semantic gap % closes the gap 
between discriminative and generative learning using % through a 
paired codebook learning.
% strategy. 
% By combining a masked token reconstruction objective with a novel codebook contrast objective, LEASE learns a shared latent space that aligns subtle and detailed information required for generation with the strong semantic representations needed for discriminative tasks. The proposed training objective does not require data augmentations or heavy online teachers, enabling an efficient training endorsed by precomputed token inputs. 
By combining masked token reconstruction with a novel codebook contrast objective, LEASE learns a shared latent space that captures both fine-grained generative detail and high-level discriminative semantics without data augmentations, online tokenizers, or frozen teachers.
Extensive experiments show that LEASE achieves strong performance across both representation learning and visual generation tasks, outperforming prior MIM and VQGAN-based unified SSL methods while competing with, or even surpassing, discriminative and generative specialized models. 
Its conditional generation results further highlight the versatility of the learned representations.
% Additionally, we further explored LEASE's features on conditional generation, demonstrating its potential to extend to guided tasks.
LEASE demonstrates the value of jointly learning from generative and discriminative semantic dictionaries, offering a promising foundation for future general-purpose unified vision models.
% We believe that LEASE reinforces the importance of generative-discriminative semantic understanding and its complementary dictionary learning provides a promising foundation for future research on general vision models 
% that jointly understand and generate visual content using the same latent space.

\section*{Acknowledgments}

\hyphenation{Deep-Sense}
\newcommand{\museNo}{01070421}

\newcommand{\DFVolNo}{PDC\-2022-133642-I00}

This work was partially funded by the 
% EU project MUSAE (No. \museNo), 
2021-SGR-01094 (AGAUR), 
Icrea Academia'2022 (Generalitat de Catalunya), 
% Robo STEAM (2022-1-BG01-KA220-VET-000089434, Erasmus+ EU), 
% DeepSense (ACE053/22/000029, ACCIÓ), 
% DeepFoodVol (AEI-MICINN, \DFVolNo) 
EXPLORA-SCALE (AIA2025-163919-C51 funded by MICIU/AEI/10.13039/501100011033), 
and 
IDEATE (PID2022-141566NB-I00, AEI-MICINN).
J. M. Rodríguez-de-Vera and Imanol G. Estepa acknowledge the support of FPU Becas with code FPU22/03116 and FPU23/02822 respectively, Ministry of Universities, Spain.
% B. Nagarajan acknowledges the support of BSC AI4Science (AI4S) Fellowship.
B. Nagarajan acknowledges AI4S fellowship within the “Generación D” initiative by Red.es, Ministerio para la Transformación Digital y de la Función Pública, for talent attraction (C005/24-ED CV1), funded by NextGenerationEU through PRTR.
The authors thankfully acknowledge EuroHPC Joint Undertaking (EHPC-AIF-2025SC01-047) 
% for awarding us access to Leonardo at CINECA, Italy 
and Spanish Supercomputing Network (RES) (IM-2025-1-0023, IM-2025-2-0045, IM-2025-3-0037) for awarding us access to MareNostrum5 at BSC, Spain.

% \clearpage

{
    \small
    \bibliographystyle{ieeenat_fullname}
    \bibliography{main}
}

\clearpage
\setcounter{page}{1}
\renewcommand{\thepage}{\roman{page}}
\setcounter{figure}{0}
\renewcommand\thefigure{\Alph{section}.\arabic{figure}}
\setcounter{table}{0} 
\renewcommand\thetable{\Alph{section}.\arabic{table}}
\setcounter{section}{0}
\renewcommand{\thesection}{\Alph{section}}
\maketitlesupplementary

The supplementary material provides additional setup and hyperparameter details, extended transfer learning results, deeper analyses of the generative and discriminative codebooks, dense-task evaluations, and qualitative generation results.

\section{Experiment setup}
For both generative and discriminative  tasks, we follow the experimental setup as in MaskGIT \cite{chang_maskgit_2022}, MAGE \cite{li_mage_2023} and Sorcen \cite{estepa_conjuring_2025}. 
In tables  \ref{tab:pretraining_settings} to \ref{tab:conditional_ft}, we show the most relevant hyperparameters for all experiments. 
We provide the robustness dataset details in \Cref{tab:datasets_corruptions}. 
Note that, for conditional generation setup (\Cref{tab:conditional_ft}), we do \textit{not} retrain the decoder, but fine-tune it exclusively on masked token reconstruction tasks. 
For guidance, we concatenate the extracted discriminative centroids of every patch and the CLIP class embedding to the main reconstruction canvas that is later fed to the decoder. 
The encoder and rest of the elements in the architecture is kept frozen during this process. 
For all pretraining, the training dataset is precomputed before the training. This process can be done in $\sim$7 hours on a single H100, including the k-Means computation required for the discriminative codebook creation. 
Note that this computation is done only once and can be considered negligible when compared against the  $\sim$951 hours introduced by the online VQGAN tokenizer used in MAGE \cite{he_masked_2022}. 

\begin{table}[!htpb]
\centering
\begin{tabular}{|c|c|}
\hline
\textbf{config} & \textbf{value} \\
\hline
optimizer & AdamW  \\
base learning rate & 1.5e-4 \\
weight decay & 0.05 \\
optimizer momentum & $\beta_1$, $\beta_2$ = 0.9, 0.95 \\
batch size & 4096\\
learning rate schedule & cosine decay  \\
warmup epochs & 40 \\
training epochs & 1600 \\
gradient clip & 3.0 \\
label smoothing & 0.1 \\
dropout & 0.5 \\
masking ratio min & 0.5  \\
masking ratio max & 1.0  \\
masking ratio mode & 0.55 \\
masking ratio std & 0.25 \\
$\lambda$ & 0.1 \\
NN number & 5 \\
$\tau$ & 0.1 \\
$\alpha$ & 0.1 \\
\hline
\end{tabular}
\caption{Pre-training Settings.}\label{tab:pretraining_settings}
\end{table}

\begin{table}[!htpb]
\centering
\begin{tabular}{|c|c|}
\hline
\textbf{config} & \textbf{value} \\
\hline
optimizer & LARS \\
base learning rate & 0.1 \\
weight decay & 0 \\
optimizer momentum & 0.9 \\
batch size & 4096 \\
learning rate schedule & cosine decay \\
warmup epochs & 0 \\
training epochs & 90 \\
augmentation & RandomResizedCrop \\
\hline
\end{tabular}
\caption{Linear Probing Settings.}\label{tab:linear_probing_settings}
\end{table}

\begin{table}[!htpb]
\centering
\begin{tabular}{|c|c|}
\hline
\textbf{config} & \textbf{value} \\
\hline
optimizer & AdamW \\
base learning rate & 2.5e-4 \\
weight decay & 0.05 \\
optimizer momentum & $\beta_1, \beta_2 = 0.9, 0.999$ \\
layer-wise lr decay & 0.65 \\
batch size & 1024 \\
learning rate schedule & cosine decay \\
warmup epochs & 5 \\
training epochs & 100  \\
label smoothing & 0.1 \\
augmentation & RandAug (9, 0.5) \\
mixup & 0.8 \\
cutmix & 1.0 \\
random erase & 0 \\
drop path & 0.1  \\
\hline
\end{tabular}
\caption{End-to-End Finetuning Settings.}\label{tab:ft_settings}
\end{table}

\begin{table}[!htpb]
\centering
\begin{tabular}{|c|c|}
\hline
\textbf{config} & \textbf{value} \\
\hline
optimizer & LARS \\
base learning rate & 1.0 \\
weight decay & 0.0 \\
optimizer momentum & 0.9 \\
batch size & 16 \\
learning rate schedule & cosine decay \\
warmup epochs & 0 \\
training epochs & 10 \\
augmentation & RandomResizedCrop \\
\hline
\end{tabular}
\caption{Few-shot Settings.}\label{tab:few_shot_settings}
\end{table}

\begin{table}[!htpb]
\centering
\begin{tabular}{|c|c|}
\hline
\textbf{config} & \textbf{value} \\
\hline
optimizer & AdamW  \\
base learning rate & 1.5e-4 \\
weight decay & 0.05 \\
optimizer momentum & $\beta_1$, $\beta_2$ = 0.9, 0.95 \\
batch size & 4096\\
learning rate schedule & cosine decay  \\
warmup epochs & 5 \\
training epochs & 300 \\
gradient clip & 3.0 \\
label smoothing & 0.1 \\
dropout & 0.5 \\
masking ratio min & 0.5  \\
masking ratio max & 1.0  \\
masking ratio mode & 0.55 \\
masking ratio std & 0.25 \\
\hline
\end{tabular}
\caption{Conditional Finetuning Settings. Note that only the decoder is finetuned.}\label{tab:conditional_ft}
\end{table}

\begin{table}[h!]
\centering
\begin{tabular}{|c|c|}
\hline
\textbf{dataset} & \textbf{corruption / specialty} \\ 
\hline
ImageNet-1k (val) & clean validation \\
ImageNet-V2 (INv2) & matched-frequency test \\ 
ImageNet-Sketch (IN-S) & sketch-based domain shift \\
ImageNet-Rend. (IN-R) & artistic renditions  \\
ImageNet-A (IN-A) & adversarial examples \\
ObjectNet (ObjN.) & object distribution shift \\
ImageNet-C & multi-corruption \\
\bottomrule
\end{tabular}
\caption{Datasets and their corresponding corruption types or domain specializations.}
\label{tab:datasets_corruptions}
\end{table}

\section{Hyperparameter Ablations}
We ablate the two main hyperparameters of the codebook contrast objective: the weighting factor $\lambda$ and the number of neighbor centroids. These ablations are performed on IN200 subsample dataset and trained for 200 epochs. As shown in \Cref{weight} $\lambda = 0.1$ yields the best results, concurring with previous works \cite{huang_contrastive_2024, li_mage_2023, estepa_conjuring_2025}. For the number of neighbor centroids, we observe that using 5 or 30 neighbors yields the strongest generative performance. However, the larger set of 30 neighbors reduces the model’s discriminative capacity. Overall, LEASE remains robust across a broad range of neighbor counts, with 5 neighbors offering the best balance, consistent with findings from prior neighbor-based contrastive frameworks \cite{koohpayegani_mean_2021, estepa_all4one_2023}.

\begin{table}[t!]
\centering
\small
\begin{subtable}[t]{0.5\linewidth}
\centering
\setlength{\tabcolsep}{2pt}
\begin{tabular}{@{}cccc@{}}
\toprule
$\lambda$ value & LP & FID & IS \\ \midrule
0.1         &  85.47  &  16.33   &  58.11  \\
0.5         &  85.43  &  17.31   &  56.31  \\
1.0         &  84.93  &  17.56   &  56.21  \\ \bottomrule
\end{tabular}
\caption{Contrastive objective weight.}\label{weight}
\end{subtable}
\hfill
\begin{subtable}[t]{0.47\linewidth}
\centering
\setlength{\tabcolsep}{2pt}
\begin{tabular}{@{}cccc@{}}
\toprule
\# NN & LP & FID & IS \\ \midrule
5          &  85.47  &  16.33   &   58.11 \\
15         &  85.50  &  16.80   &  57.25  \\
30         &  85.36  &   16.31  &  58.29  \\ \bottomrule
\end{tabular}
\caption{Number of neighbors ablation.}\label{NN_number}
\end{subtable}
\caption{Ablation results for linear probing accuracy (LP), FID and IS.} %: (a) contrastive objective weight and (b) number of neighbors.}
\end{table}

\section{Extended Transfer Learning Results}
In \Cref{tranfer8} and \Cref{tranfer4}, we extend transfer learning experimentation to 8-shot and 4-shot regimes. In the 8-shot scenario, LEASE demonstrates the strongest overall performance, achieving the highest average accuracy across datasets and outperforming previous VQGAN-based methods \cite{li_mage_2023, estepa_conjuring_2025} in 6 out of 7 datasets. In the 4-shot setting LEASE maintains a clear advantage, showing the highest average accuracy among the three methods and outperforms both competitors on all datasets. 
As shown in 16-shot setting (Refer \Cref{tab:transfer16} from the main manuscript), LEASE consistently shows superior few-shot transfer performance, while 8-shot and 4-shot settings prove that this margin increases as the number of shots decreases. 

\begin{table}[t!]
\small
\centering
\setlength{\tabcolsep}{2pt}
\begin{tabular}{@{}lccccccccc@{}}
\toprule
              & Caltech & UCF101 & Sun   & DTD   & C100 & C10 & Places & Avg.        \\ \midrule
MAGE          & 83.94   & 46.37  & 43.38 & 40.13  & 53.08        & \textbf{75.02}       & 27.00         & 52.70 \\
Sorcen        & 85.31   & \textbf{50.70}  & 44.65 & 39.60 & 49.73    & 65.02   & 27.32     & 51.76 \\
LEASE         & \textbf{87.06}   & \textbf{50.91}  & \textbf{47.86} & \textbf{42.02} & \textbf{56.60}    & 72.99   & \textbf{31.12}     & \textbf{55.51}       \\ \bottomrule
\end{tabular}
\caption{Transfer learning results (Top-1 accuracy (\%)) for different datasets under 8-shot settings. % and average across datasets.
}\label{tranfer8}
\end{table}

\begin{table}[t!]
\small
\centering
\setlength{\tabcolsep}{2pt}
\begin{tabular}{@{}lccccccccc@{}}
\toprule
              & Caltech & UCF101 & Sun   & DTD   & C100 & C10 & Places & Avg.        \\ \midrule
MAGE          & 71.72   & 29.71  & 31.36 & 19.92  & 40.03        & 35.06       & 19.81         & 35.37 \\
Sorcen        & 69.20   & 36.35  & 33.48 & 21.99 & 35.58    & 34.44   & 20.32     & 35.91 \\
LEASE         & \textbf{78.22}   & \textbf{39.62}  & \textbf{36.77} & \textbf{27.01} & \textbf{47.42}    & \textbf{44.04}   & \textbf{24.72}     & \textbf{42.54}       \\ \bottomrule
\end{tabular}
\caption{Transfer learning results (Top-1 accuracy (\%)) for different datasets under 4-shot settings. % and average across datasets.
}\label{tranfer4}
\end{table}

\section{Generative vs. Discriminative Codebook}

% \paragraph{Codebook Pair Relationship.}
\paragraph{Relationship between Codebook Pairs.}

\begin{figure}[t!]
    \centering
    \includegraphics[width=0.49\textwidth]{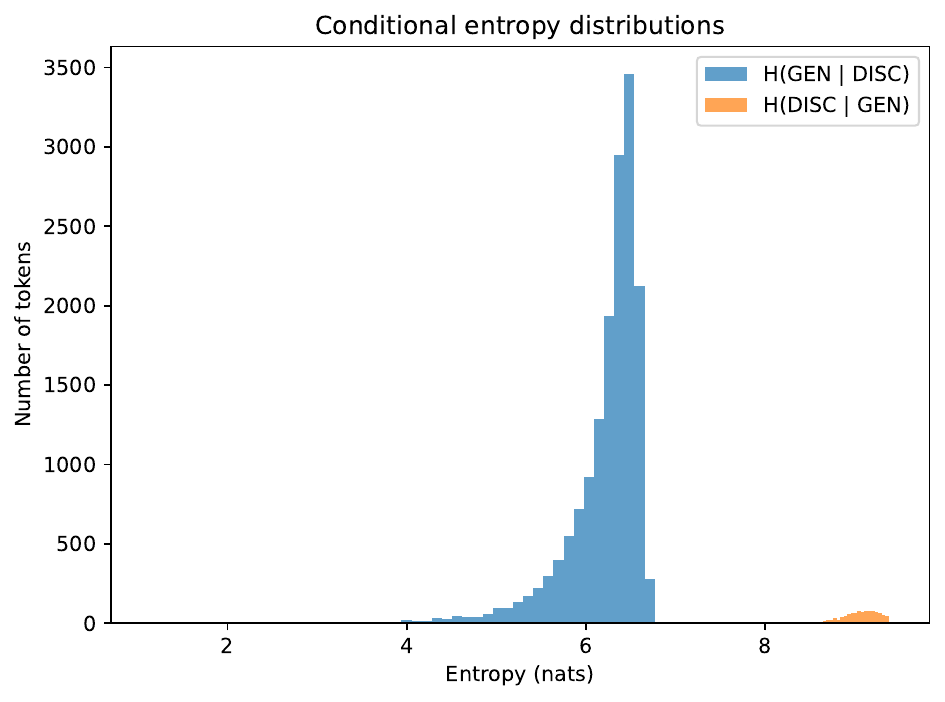}
    \caption{Conditional entropy distributions for discriminative and generative codebooks. % comparison for 1600 epoch of VQGAN-based unified methods
    }
    \label{fig:cod_vs_1}
\end{figure}

We analyse how generative (GEN) and discriminative (DISC) codebooks relate by computing the conditional entropies $H(GEN|DISC)$ and $H(DISC|GEN)$ from the patch-aligned token co-occurrence matrix. This conditional entropy $H(X|Y)$ measures how uncertain the value of a random variable $X$ remains when another variable $Y$ is known. In practice, it quantifies the average amount of information required to describe $X$ after knowing $Y$. Low conditional entropy means knowing
$Y$ makes $X$ highly predictable. \Cref{fig:cod_vs_1} shows the distribution of these entropies across all tokens (measured in nats), and reveals a strong asymmetry between the two directions.

A subset of discriminative tokens leads to low $H(GEN|DISC)$, meaning that these discriminative tokens consistently co-occur with a small set of generative tokens. In contrast, $H(DISC|GEN)$ is uniformly high, indicating that generative tokens provide almost no information about the corresponding discriminative tokens. This reflects the fundamental difference between the two codebooks: the generative codebook, based on VQGAN, is texture-based and largely class-agnostic, whereas the discriminative codebook, made by the features of DINOv2, encodes semantic distinctions learned through local appearance signals. Consequently, even if discriminative tokens do not explicitly encode textures, their semantic prototypes are tied to characteristic visual patterns that allow generative tokens to be predicted in some cases. This observation is consistent with prior work such as DiGIT \cite{zhu_stabilize_2024}, which is able to train a generative model solely from a codebook based on DINOv2. Still, the low-entropy behaviour only occurs for a small amount of discriminative tokens ($\sim$500 out of 16K), while the majority produce entropies above 6 nats. 
This indicates that the semantic overlap between the two codebooks is limited: most discriminative tokens correspond to a broad set of VQGAN textures, and generative tokens do not encode the semantic distinctions present in the discriminative codebook.

\paragraph{Class-average Token Distributions.}

\begin{figure}[t]
    \centering
    \includegraphics[width=0.49\textwidth]{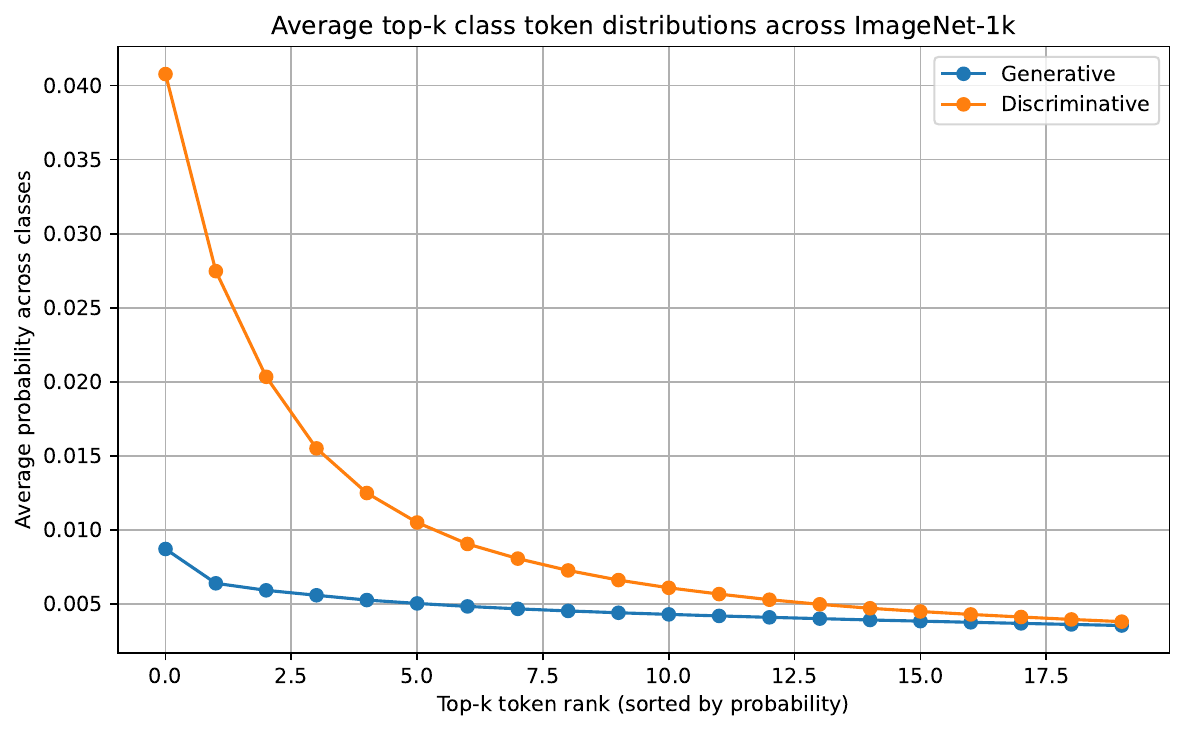}
    \caption{Average top-k token distributions across all ImageNet classes.}
    \label{fig:cod_vs_2}
\end{figure}

We display the semantical contrast between the generative and discriminative codebooks used to train LEASE by computing the top-k token probabilities for each ImageNet class and averaging them across all 1000 classes. In practice, we extract the frequency of every encoded patch token on all ImageNet classes for both codebooks. Most frequent tokens for a given class are assumed to contain more discriminative information about that specific class. At the same time, the higher the frequency the higher the discriminative information, as it would mean that an specific class is better represented by the token. These frequencies are represented as a probabilities for every class on ImageNet and averaged. In \Cref{fig:cod_vs_2}, we can see the average probability of the top 20 most probable tokens per class. 

The discriminative codebook exhibits a sharp, heavy-tailed distribution, where a small number of tokens dominate the class representation. This indicates strong class-specific semantics. In contrast, the generative codebook produces an almost flat curve, with top tokens only marginally more probable than lower ones. This marginality, combined with the masked token reconstruction strategies, could explain the discriminative capacity of works such as MAGE \cite{li_mage_2023} and Sorcen \cite{estepa_conjuring_2025}, which leverage a generative codebook exclusively. Still, this behavior confirms that generative tokens carry little class-specific information and mainly encode generic textures. The aggregation over all classes demonstrates that this behavior is global rather than class-specific. The discriminative codebook, even if it comes from a self-supervised model, consistently organizes patches into semantically meaningful groups, while the generative codebook remains largely class-agnostic.

\paragraph{Token-level Shared Structure.} 

\begin{figure}[t]
    \centering
    \includegraphics[width=0.49\textwidth]{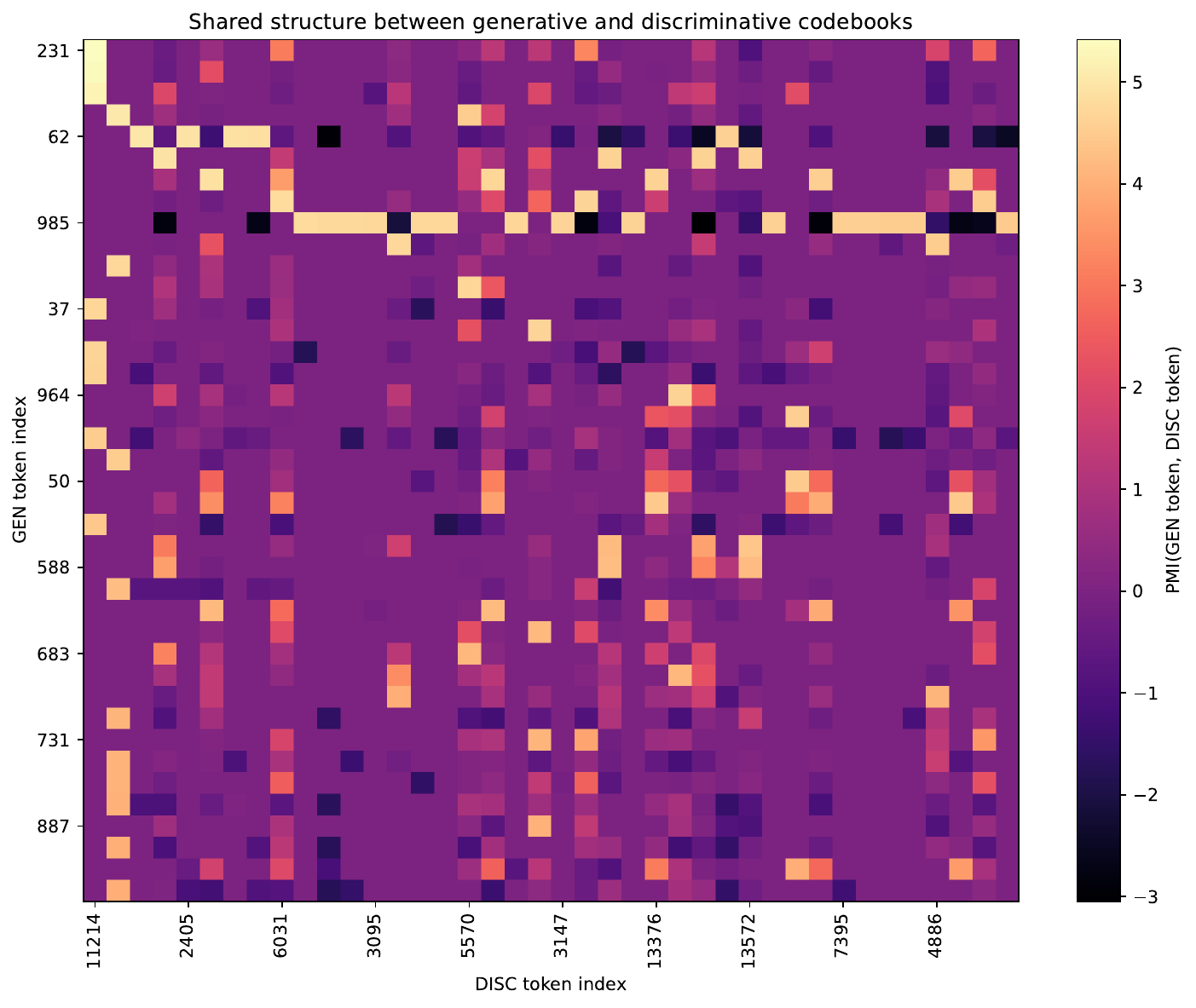}
    \caption{PMI heatmap between the generative and discriminative codebooks.
    }
    \label{fig:cod_vs_3}
\end{figure}

Finally, to visualize the overlap between the two codebooks, we compute the pointwise mutual information (PMI) between every generative and discriminative token and display a heatmap for the pairs with highest PMIs (40 tokens per axis in total) in \Cref{fig:cod_vs_3}. Each column corresponds to one discriminative token and each row to one generative token. We observe only a few isolated bright cells per column, indicating that a small number of discriminative tokens consistently co-occur with some specific generative tokens. The majority of entries are close to zero PMI, even in this top-PMI subset, and we do not see large block structures that would indicate a shared latent organization. This confirms that the overlap between the generative and discriminative codebooks is sparse. Individually, both codebooks contain useful information for their respective domains, either generation or discrimination. However, these semantics are not shared among both. The limited overlap is not simply an effect of vocabulary size, but rather reflects a fundamental difference in what the two codebooks represent: the discriminative codebook organizes patches according to semantic object structure, while the generative codebook captures texture-based, class-agnostic appearance. 

%Consequently, some discriminative tokens have characteristic texture signatures captured by generative codebook, but most of the semantic structure in the discriminative codebook has no counterpart in the generative one.

\section{Dense Task Evaluation}

Token-based approaches, including MAGE \cite{li_mage_2023} and Sorcen \cite{estepa_conjuring_2025}, continue to face limitations on dense prediction tasks due to the loss of fine-grained spatial information \cite{tian_addp_2024} caused by the quantized input tokens. As shown in \Cref{dense}, all methods achieve comparable performance on MSCOCO, with Sorcen and LEASE slightly improving over MAGE. On FoodSeg, a more specific dataset, Sorcen stands as the best while LEASE matches the performance obtained by MAGE. 

\renewcommand{\arraystretch}{0.85}
\begin{table}[h!]
\centering
\begin{tabular}{@{}llll@{}}
\toprule
       & MSCOCO  & FoodSeg \\ \midrule
MAGE   & 15.80  & 15.88   \\
Sorcen & 15.90  & 16.82   \\ 
LEASE &  15.92 & 15.85 \\ \bottomrule
\end{tabular}
\caption{Extended evaluation on Instance Segmentation downstream task. mAP metric is reported.}\label{dense}
\end{table}

\FloatBarrier

\section{Generation Visualization}
\begin{figure*}[h]
    \centering
    \includegraphics[width=0.90\textwidth]{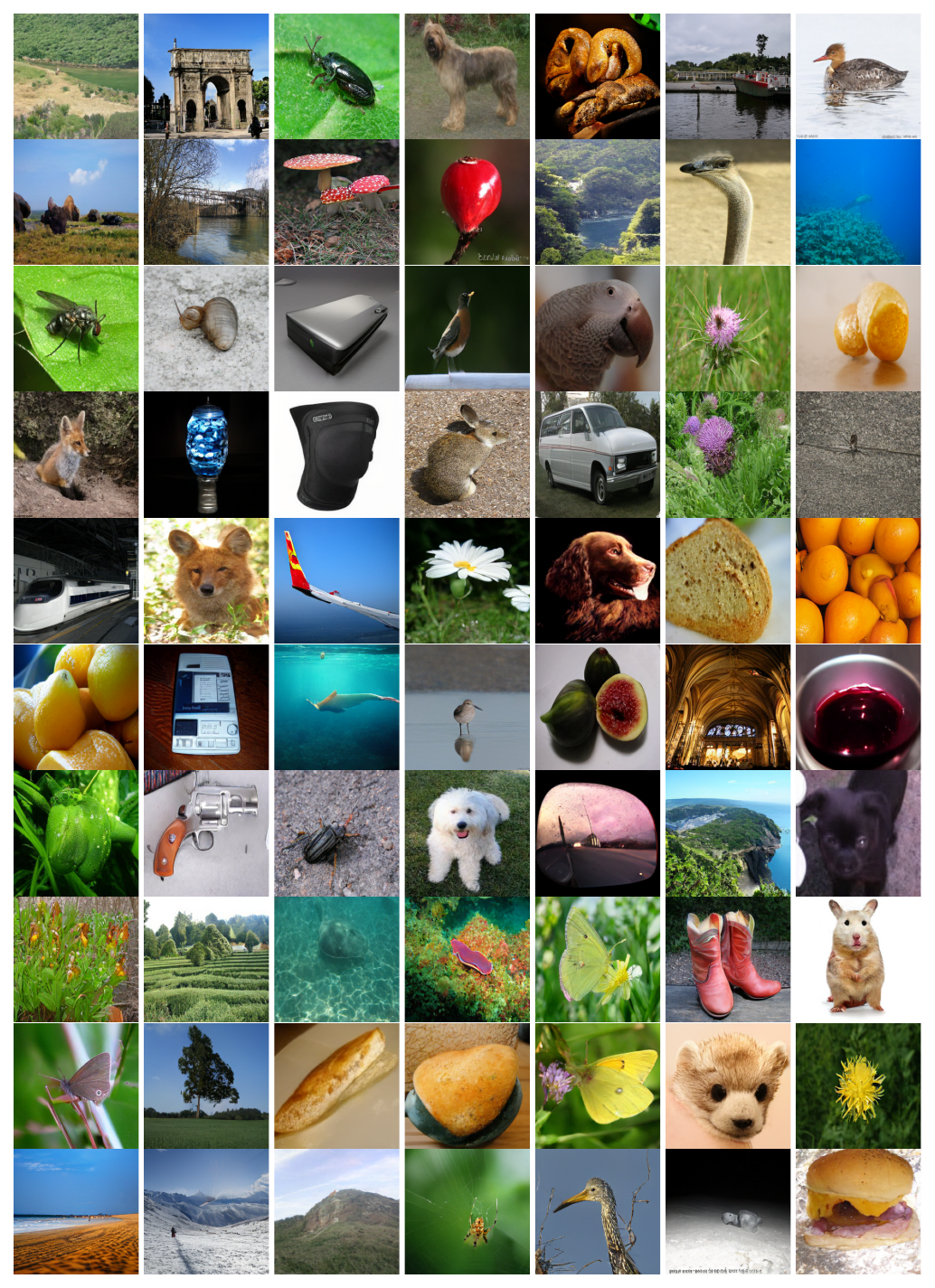}
    \caption{Unconditional generation examples of LEASE using ViT-B. % comparison for 1600 epoch of VQGAN-based unified methods
    }
    \label{fig:uncond}
\end{figure*}

\begin{figure*}[h]
    \centering
    \includegraphics[width=0.90\textwidth]{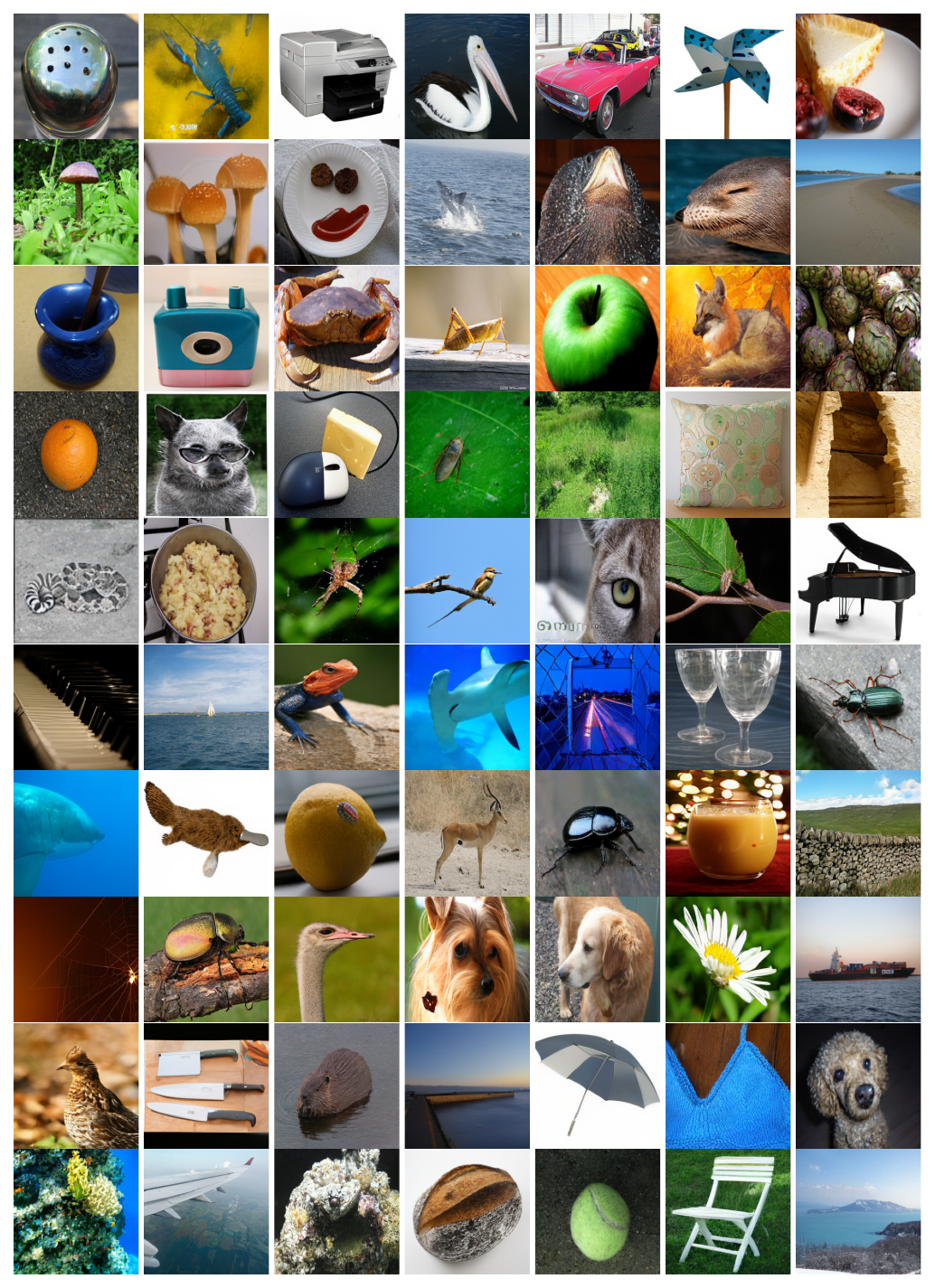}
    \caption{Conditional generation examples of LEASE using ViT-B. % comparison for 1600 epoch of VQGAN-based unified methods
    }
    \label{fig:cond}
\end{figure*}

\begin{figure*}[h]
    \centering
    \includegraphics[width=0.80\textwidth]{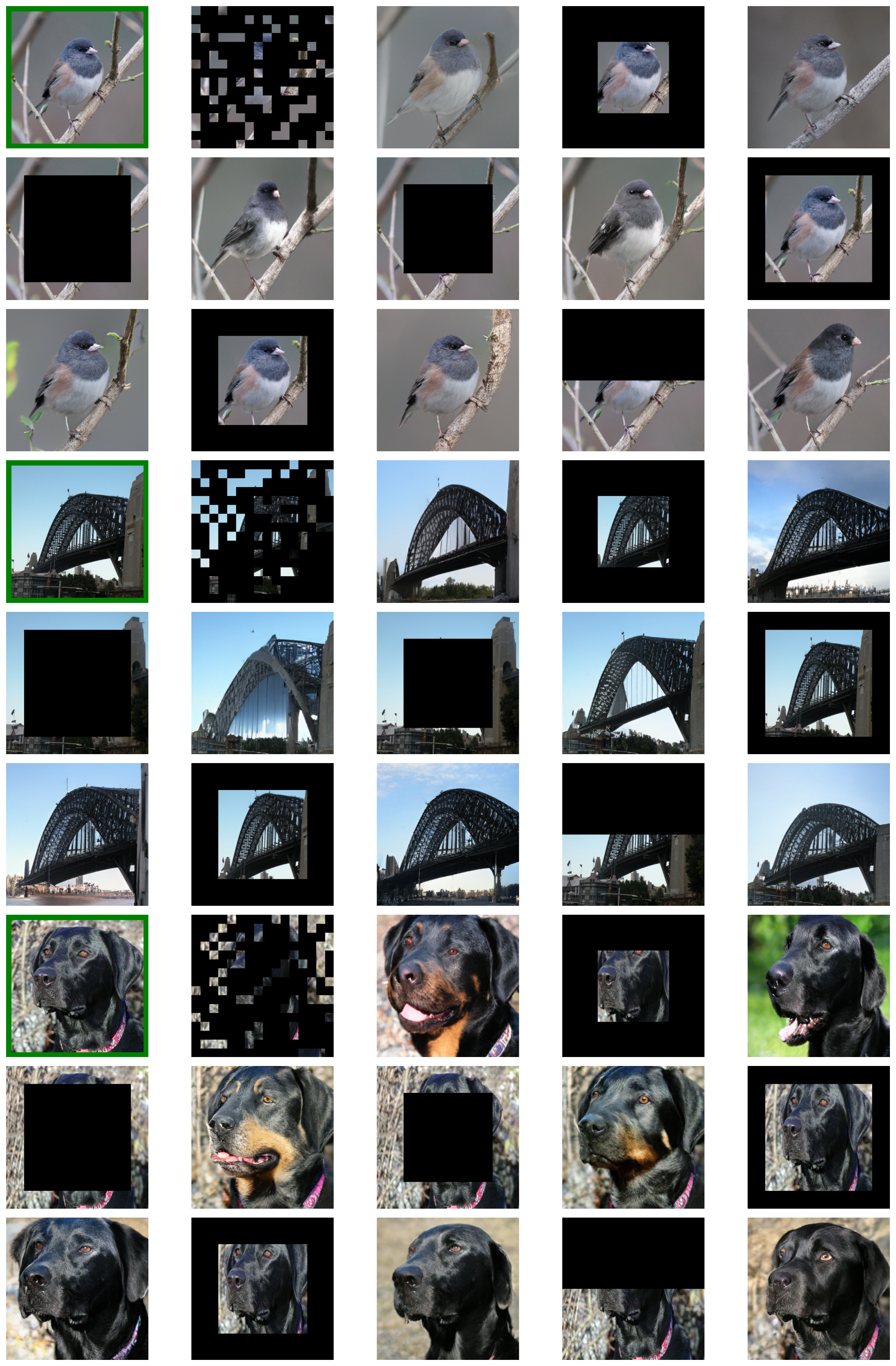}
    \caption{Examples of image reconstruction, inpainting and outpainting for LEASE using ViT-B. Original image is marked in green. % comparison for 1600 epoch of VQGAN-based unified methods
    }
    \label{fig:recon1}
\end{figure*}

\begin{figure*}[h]
    \centering
    \includegraphics[width=0.80\textwidth]{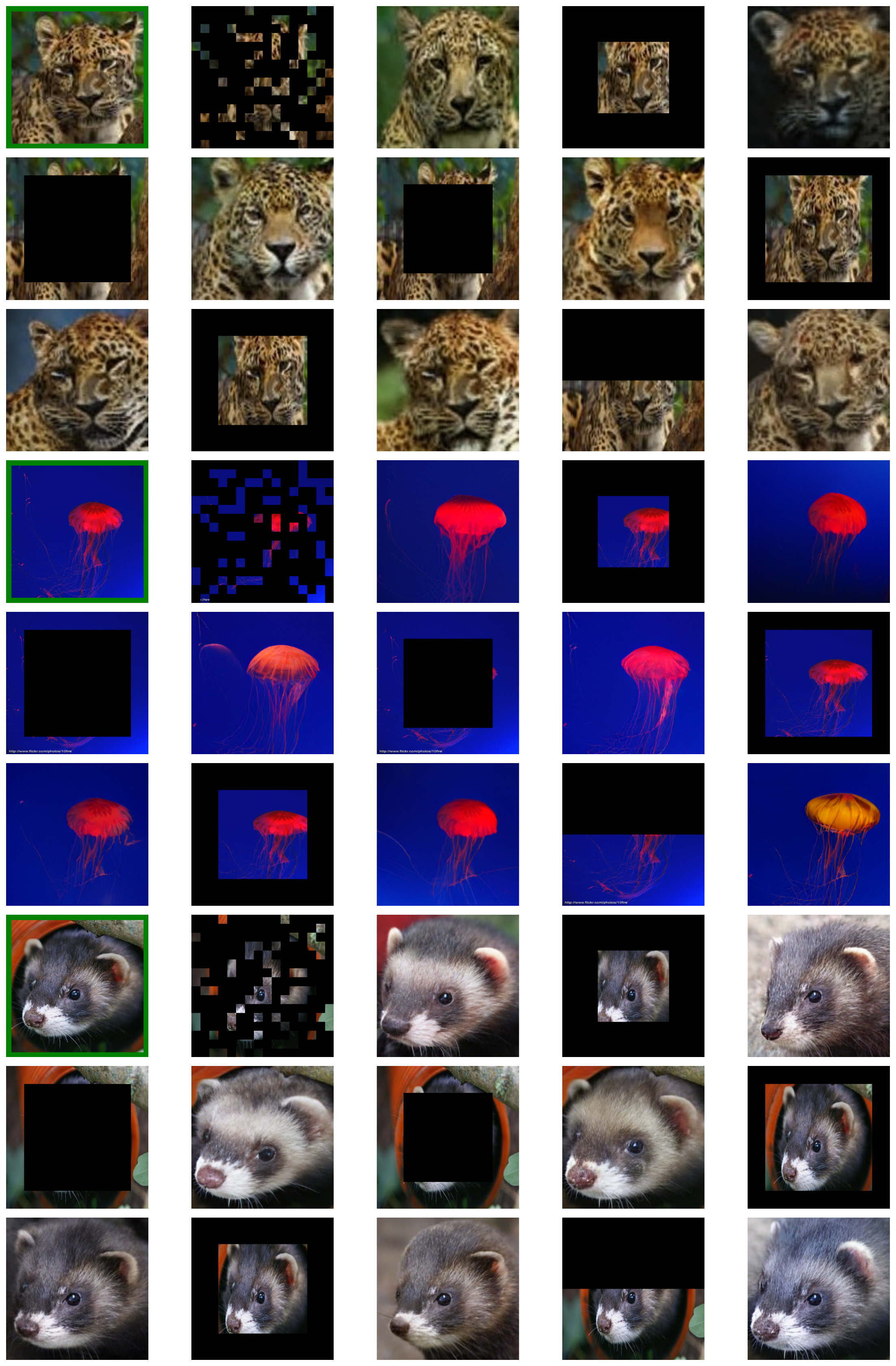}
    \caption{Examples of image reconstruction, inpainting and outpainting for LEASE using ViT-B. Original image is marked in green. % comparison for 1600 epoch of VQGAN-based unified methods
    }
    \label{fig:recon2}
\end{figure*}

To further illustrate the generative capabilities of LEASE, we present qualitative samples under both unconditional and class-conditional settings in \Cref{fig:uncond} and \Cref{fig:cond}. Additionally, we also visualize the reconstruction, inpainting and outpainting capacity of LEASE in \Cref{fig:recon1} and \Cref{fig:recon2}.

\end{document}